\documentclass{kais}
\usepackage{hyperref}
\usepackage{wrapfig,epsf}
\usepackage[dcucite]{harvard}
\usepackage{amssymb}
\usepackage{graphicx}
\usepackage{listings}
\usepackage{color}
\usepackage{hhline}
\usepackage{multirow}
\usepackage{algorithm}
\usepackage{algorithmic}

\usepackage[T1]{fontenc}
\usepackage{lmodern}

\newcommand{\cost}{cost}

\newcommand{\openGoals}{openGoals}

\newcommand{\G}{{\mathcal G}}

\newcommand{\D}{{\mathcal D}}
\newcommand{\F}{{\mathcal F}}
\newcommand{\I}{{\mathcal I}}
\newcommand{\A}{{\mathcal A}}
\newcommand{\T}{{\mathcal T}}
\newcommand{\AG}{{\mathcal{AG}}}
\newcommand{\PG}{{\mathcal{PG}}}
\newcommand{\Prop}{{\mathcal V}}

\newcommand{\CL}{{\mathcal{CL}}}

\newcommand{\Obj}{{\mathcal{O}}}

\newcommand{\Or}{{\mathcal{OR}}}

\newtheorem{definition}{Definition}

\received{Jan 25, 2012}
\revised{Jun 15, 2012}
\accepted{Sep 01, 2012}

\pubyear{2011}
\pagerange{\pageref{firstpage}--\pageref{lastpage}}
\volume{xxx}

\begin{document}
\label{firstpage}

%\title{A Flexible Coupling Approach to Cooperative Planning for Multiple Agents under Incomplete Information}
\title{A Flexible Coupling Approach to Multi-Agent Planning under Incomplete Information}

\author[A. Torre\~no et al]{Alejandro Torre\~no $^1$, Eva Onaindia$^1$ and \'Oscar Sapena$^1$
\\ $^1$Departamento de Sistemas Inform\'aticos y Computaci\'on, Universitat Polit\`ecnica de Val\`encia, \\Valencia, Spain}

\maketitle

\begin{abstract}
Multi-agent planning (MAP) approaches are typically oriented at solving loosely-coupled problems, being ineffective to deal with more complex, strongly-related problems. In most cases, agents work under complete information, building complete knowledge bases. The present article introduces a general-purpose MAP framework designed to tackle problems of any coupling levels under incomplete information. Agents in our MAP model are partially unaware of the information managed by the rest of agents and share only the critical information that affects other agents, thus maintaining a distributed vision of the task.

%Multi-agent planning (MAP) approaches are typically oriented at solving loosely-coupled problems, which require little coordination between the agents' sub-plans. However, when it comes to more complex, strongly-related problems, MAP has been traditionally discarded in favour of centralized models. The present article introduces a general-purpose MAP framework designed to tackle problems of any coupling levels under incomplete information. Agents in our MAP model are partially unaware of the information managed by the rest of agents and share only the critical information that affects other agents, thus maintaining a distributed vision of the task.

Agents solve MAP tasks through the adoption of an iterative refinement planning procedure that uses single-agent planning technology. In particular, agents will devise refinements through the Partial-Order Planning paradigm, a flexible framework to build refinement plans leaving unsolved details that will be gradually completed by means of new refinements. Our proposal is supported with the implementation of a fully-operative MAP system and we show various experiments when running our system over different types of MAP problems, from the most strongly-related to the most loosely-coupled.\end{abstract}


\section{Introduction}
\label{Intro}
\definecolor{lightgray}{gray}{0.92}
\definecolor{darkgreen}{rgb}{0,0.5,0}
\lstset{basicstyle=\footnotesize, showtabs=true, frame=single, backgroundcolor=\color{lightgray}, escapechar=@, emph={functions, multi-functions, action, predicates, requirements, types, domain, problem, objects, init},emphstyle={\color{blue}\bfseries}}

Planning is the art of building control algorithms that synthesize a course of action to achieve a desired set of goals from an initial situation. Traditionally, planning has been regarded as a centralized process in which a single entity is in charge of devising a plan that satisfies the problem goals.

Multi-Agent Planning (MAP) generalizes the problem of planning in domains where several agents plan and act together. MAP introduces a social approach to planning \cite{Nguyen09}, focusing on the collective effort of multiple planning entities to accomplish tasks by combining their knowledge, information and capabilities. This is required when agents are unable to solve their tasks by themselves, or at least can accomplish them better (more quickly, completely, precisely, or certainly) when working with others \cite{Durfee01}.

MAP is concerned with planning \emph{by} multiple agents, i.e., distributed planning, and planning \emph{for} multiple agents, i.e., planning for multi-agent execution, thus giving rise to a great variety of tools and techniques. The approach traditionally adopted by the Multi-Agent Systems (MAS) research community assumes that, in general, agents are self-interested and that there is not a common goal to solve, thus focusing on coordinating the activities of multiple agents in a shared environment \cite{Desjardins99a}. In agent-oriented approaches, the ultimate objective is to ensure that the agents' local objectives (private goals) will be achieved by their plans and so the emphasis is put on distributed execution, plan synchronization and collaborative activity at run-time planning \cite{Durfee91,Tambe97,KaminkaPT02}. All in all, these techniques use planning as a means to controlling and coordinating agents rather than building a competent and joint plan, and so they are very appropriate for the design of real-time systems \cite{Micacchi08}.

In planning-oriented approaches dealing with contexts in which agents are assumed to be cooperative, the objective is to study how planning can be extended into a distributed environment or, more particularly, on the construction of a competent plan by several planning entities. There exist different approaches to address this objective, varying according to the typology of the planning problem to solve. In particular, the adoption of one or another strategy depends on the coordination needs of the problem, i.e., to which extent agents are able to make their own plans without affecting what the other agents are planning to do. Thus, when agents are assumed to be relatively independent, they carry out their planning activities individually and exchange information about their local plans, which they iteratively refine and revise until they fit together in order to ensure that the resulting plan will jointly execute in a coherent and efficient manner \cite{Desjardins99a}. This has been the predominant approach in cooperative MAP, existing a large body of research on post-planning coordination, i.e., solving inconsistencies among local plans that have been constructed separately. The well-known Partial Global Planning (PGP) framework (Durfee and Lesser, 1991) is one of the first techniques that allows agents to communicate and merge their local plans. Ever since, many works on plan merging methods for building a joint plan given the local plans of each participating agent have arisen (see section \ref{Background} for a detailed description).

The application of MAP to loosely-coupled multi-agent tasks, in which agents have little interaction to each other, is still an active area of research. Some recent works in this line, where agents are engaged in some cooperative behaviour, have emerged lately. These works follow a common approach that consists of coordinating the local solutions developed by the agents. For instance, the work in \cite{Kvarnstrom11} considers that agents have sequential threads of execution and interactions only occur when distributing sub-plans to individual agents for plan execution. This approximation follows a single-agent planning and distributed coordination. The work in \cite{Brafman08} applies individual planning and coordinates the local solutions through the resolution of a Constraint Satisfaction Problem (CSP). In an extension of this latter work, authors use a distributed CSP to solve inconsistencies among agents' plans \cite{Nissim10}.

Most of the aforementioned approaches turn out to be inefficient at the time of solving strongly-related problems in which the number of coordination points among agents is large \cite{Nissim10}. To deal with these problems, other MAP models use a unified approach in which planning and coordination of activities are integrated rather than being treated as independent processes \cite{JonssonR11,BelesiotisRR10}. However, these approaches do not achieve high performance in loosely-coupled problems because the reasoning procedures rely very strongly on a high degree of interdependency between the agents' actions.

The problem of building a competent joint plan among several planning entities has been generally dismissed by the MAS community, more concerned with the development of coordination mechanisms for agents, and ignored by the planning community, which has traditionally resorted to efficient single-agent algorithms to solve planning problems. MAP is not only about a divide-and-conquer strategy to tackle large planning problems, it is also about the development of techniques for planning entities that are geographically or spatially distributed. While one might expect the number of coordination points in inherently distributed problems not to be very large, another issue that comes up is the distribution of information among agents. In frameworks like those presented in \cite{Brenner09,BelesiotisRR10} agents communicate all the available information and build complete knowledge bases, i.e., agents have complete information on the MAP task. However, in large-size problems with heterogeneous agents, building complete knowledge bases is not viable. Besides efficiency issues, agents may be unable to manage the information handled by other agents as they may have different knowledge and abilities.

In this paper, we present a novel approach to cooperative MAP that allows to efficiently solve problems with any level of interaction among agents. Unlike other techniques, our MAP system is capable of solving from the most loosely-coupled problems to the most strongly-related problems. The key point to address this aspect is to use a refinement planning approach \cite{Kambhampati97} that allows agents to interleave planning and coordination, or more specifically, to coordinate their plans during planning. We also allow heterogeneous agents to work under incomplete information, sharing only the critical information that affects other agents and maintaining a distributed vision of the MAP task. This issue, which has been ignored in almost all of the MAP approaches, is of key importance to efficiently handle inherently distributed problems. Last but not least, our MAP approach is entirely based on the use of single-agent planning technology adapted to a multi-agent context. More precisely, agents follow the Partial-Order Planning paradigm \cite{Nguyen01,Younes03}.

As well as introducing the MAP architecture and a theoretical model for multi-agent planning, our proposal is supported with the implementation of a fully-operative MAP system. The empirical evaluation of the system demonstrates this novel approach to be effective when dealing with both strongly-related problems and loosely-coupled problems in which agents manage incomplete information.

This paper is organized as follows: section \ref{Background} summarizes some background on the main topics related to this work and reviews the most recent literature on MAP; section \ref{M_example} introduces the example MAP scenario we will use to illustrate the different aspects of our framework; section \ref{Arch} outlines our MAP architecture; section \ref{Planning model} presents the theoretical planning model upon which our system is based; section \ref{Specification} outlines the planning language used to model MAP tasks; section \ref{Planning_reasoning} provides an overview of the MAP algorithm followed by the agents; section \ref{Initial} describes the first stage of our MAP algorithm, the initial information exchange; section \ref{Problem_solving} outlines second stage of the MAP algorithm, the refinement planning and coordination protocol; section \ref{Results} presents the experimental results, and finally, section \ref{Conclusions} concludes and summarizes our future lines of research.

%section \ref{Example} introduces an example of application that describes the modeling and resolution of a simple MAP task;

\section{Background}
\label{Background}

Our MAP model builds upon several single-agent planning techniques. This section provides a review on the principal single-agent planning concepts used in our MAP approach as well as the most relevant and recent approaches to cooperative MAP. We also outline the most relevant works on MAP architectures and frameworks and we conclude by summarizing the main contributions and novelties of our approach.

\subsection{Single-agent planning}

Single-agent planning is regarded as a search process by which a single entity synthesizes a set of actions (plan) to reach a set of objectives from an initial situation \cite{Weld99}. Over the last years, single-agent planning has experienced great advances, specifically in the construction of domain-independent heuristics. Nowadays, it is possible to find a great variety of planning systems. The most recent planners combine different techniques in order to increase the algorithms efficiency: landmarks \cite{Richter10}, domain transition graphs \cite{Helmert06}, forward-chaining partial-order planning \cite{Coles10}, probes \cite{Lipovetzky11} or divide-and-conquer strategies \cite{Dreo11}, among others.

The work in \cite{BlumF97} introduced the concept of Relaxed Planning Graph, which has proven to be one of the most effective constructs to devise heuristics in state-space planning \cite{Hoffmann01}. This technique has been integrated in many single-agent planning frameworks and has also been extended to a distributed context \cite{Feng07}.

While state-space planners such as Fast Forward \cite{Hoffmann01} are still a relevant research topic, plan-space planning has been replaced by other more efficient techniques. However, plan-space planning has recently seen a revival since its flexibility makes it specially suitable for distributed environments.

Among plan-space search algorithms, the Partial-Order Planning (POP) approach \cite{Penberthy92,Younes03} is particularly relevant. POP performs a plan-based, backward search process, refining partial plans through the addition of actions, causal links and ordering constraints. POP is based on the \emph{least commitment strategy} \cite{Weld94}, which defers planning decisions during the search process and introduces partial-order relations among actions rather than enforcing a concrete order among them. The particular nature of the POP paradigm (absence of states, backward search) makes it difficult to devise competitive heuristics to guide the search process. Although some recent works reformulate the basic algorithm to improve its performance \cite{Coles10}, POP has been discontinued by the planning community in favor of other approaches. Nevertheless, it is still used in temporal planning and MAP environments as it is a flexible paradigm to handle concurrency \cite{Boutilier01}.

\subsection{Cooperative Multi-Agent Planning}

%Multi-Agent Planning
MAP extends the single-agent planning problem by distributing the planning task among several entities which work together to devise a competent joint plan that meets the problem goals. This generalization entails some differences to the more restrictive single-agent planning approach. MAP can be viewed as the problem of coordinating agents in a shared environment where information is distributed \cite{Desjardins99a}. This definition emphasizes two aspects of MAP that are not present in single-agent planning: the coordination of the planning activities and the distribution of the information among agents.

In general, solving a cooperative MAP task involves the following stages \cite{Durfee01}: 1) global goal refinement, 2) task allocation, 3) coordination before planning, 4) individual planning, 5) coordination after planning, and 6) plan execution. Some of the previous stages can be avoided or combined. For instance, some works do not distribute the goals explicitly (avoiding stage 2) \cite{BelesiotisRR10,Brenner09}, while others apply only coordination after planning (avoiding stage 3) \cite{Krogt05b,Cox05}.

MAP problems can be classified according to their coupling level, a measure of the number of interactions or coordination points among agents that will arise during the task resolution \cite{Brafman08}. In loosely-coupled problems, each problem goal problem is likely to be solved by a single agent, while goals in strongly-related problems tend to require the cooperation of several agents. The number of coordination points in a MAP problem determines which approaches are more suitable to solve it efficiently.

A wide range of MAP approaches put the emphasis on coordination after individual planning (coordination is performed at stage 5 of the MAP scheme described above). This way, these frameworks perform the planning and coordination stages independently and separately, combining or merging solutions into a global joint plan \cite{Durfee01,deWeerdt09,ToninoBWW02,KaminkaPT02}.

Different coordination techniques have been proposed for merging and gathering several individual plans into a single joint plan. The Partial Global Planning framework \cite{Durfee91} and its extension, the Generalized Partial Global Planning approach \cite{Decker92}, allow agents to communicate their local plans to the rest of agents and then they merge this information into their own partial global plan in order to improve it. This iterative process goes on until the agents' local plans fit together. The work in \cite{ToninoBWW02} proposes a post-planning coordination approach based on the iterative revision of the agents' local plans. Agents in this model cooperate by mutually adapting their individual plans, with a focus on maximizing their common or individual profit. \cite{Nissim10} introduces a cooperative MAP approach for loosely-coupled systems in which agents carry out planning individually through a state-based planner \cite{Hoffmann01,Coles08}. The resulting local plans are then coordinated by solving a distributed Constraint Satisfaction Problem. The approach in \cite{Krogt05b} solves inconsistencies among the local plans devised by self-interested agents through plan repair. Other proposals deal with insincere agents by combining planning, coordination, and execution \cite{EphratiR96} or consider the communication needs that arise when plans are being executed \cite{Tang10}.

The aforementioned plan merging methods follow a common approach: agents build plans individually while a subsequent independent process is used to coordinate these plans. This approach is suitable for solving loosely-coupled problems efficiently as the agents' local solutions in these problems present few interdependencies with each other. Thus, plan merging through post-planning coordination is an appropriate method to tackle problems in which agents can solve the different problem goals independently and the majority of the environment resources are not shared.

However, plan merging methods present several limitations. On the one hand, goals must be a priori allocated to each agent or at least implicitly distributed among the planning entities, as agents perform their planning activity in an isolated manner. Because of this, methods based on plan merging lose flexibility against other MAP proposals. On the other hand, the previous merging approaches have proven to be inefficient when solving strongly-related problems in which most of the resources are shared and most of the goals require cooperation among agents \cite{Nissim10}. The individual planning combined with a post-planning coordination strategy is not adequate to solve these strongly-related problems, since merging may introduce exponentially many ordering constraints in problems which require a coordination effort.

Another research trend on cooperative MAP stresses the importance of combining and integrating planning and coordination activities, i.e., apply coordination during planning. Hence, this trend can be seen as an extension of single-agent planning to MAP, providing a unified vision of MAP. Proposals in this line focus on the cooperative incremental construction of a joint plan, allowing agents to perform their planning activity over a centralized plan representation. This is a more suitable approach than the plan merging techniques for tackling strongly-related MAP problems with a large number of coordination points, as agents work over a centralized plan representation and planning and coordination of activities are carried out in an integrated way.

The proposal in \cite{Desjardins99a} applies the continual planning approach, which interleaves planning and execution and coordinates agents by synchronizing them at execution time \cite{Brenner09}. The approach in \cite{JonssonR11} introduces the best-response planning algorithm, which iteratively improves the quality of the agents' plans through single-agent planning technology. Finally, the works in \cite{BelesiotisRR10,Pajares12} solve inconsistencies among agents' plans through a coordination protocol based on iterated dialogues. Agents discuss and argument about the different plan proposals until the agents' viewpoints are aligned and an agreement is reached.

The integrated planning and coordination approach followed by the aforementioned MAP models copes with a wider range of MAP problems than the plan merging method, which can only deal with simpler, loosely-coupled problems. In addition, the continual revision and coordination of the agents' plans provides better results in terms of plan quality. However, integrating planning and coordination entails higher communication costs for loosely-coupled problems than using plan merging, as coordination has to be performed throughout the planning process, thus introducing an overhead. Hence, the simpler plan merging approach is far more effective for small-size and non-complex planning tasks.

Research on cooperative MAP, traditionally carried out by the planning community, has generally overlooked the management of incomplete information, an active research topic, though, within the MAS community. Planning with incomplete information has several different meanings: that certain facts of the initial state are not known, that operators can have random or nondeterministic effects, or that the plans built contain sensing operations and are branching  \cite{HaslumJ99}. In our case, we interpret incomplete information as agents not knowing the initial state completely and being total or partially unaware of the information managed by other agents.

The issue of incomplete information has been treated from two different perspectives: the probabilistic way, with the development of formal models such as Dec-POMDPs (Decentralized Partial Observable Markov Decision Processes) for coordination among multiple agents in contexts with partial observability \cite{WuZC11,KZTijcai11}; and the epistemological way, which assumes that agents have beliefs about the state of the world and beliefs over the other agents' knowledge \cite{Kraus97a,Doshi07}. This latter approach has been widely used in games of incomplete information \cite{GmytrasiewiczD05}. Both perspectives define agents as having an imprecise or uncertain view of the world and of the other agents' information but, to the best of our knowledge, there are not proposals to deal with ignorance, i.e., local views of agents that reflect agent's unawareness over the information of the rest of agents. This introduces a complexity factor in the planning process as agents can only plan on the basis of their information, being ignorant on the planning decisions of other agents. It is important to note, though, that the information unknown to one agent does not have a direct impact on the agents' choices because its actions are not involved with the unknown piece of information. However, this absence of information may have an indirect impact in the overall planning process and quality of the plan.

\subsection{Architectures and frameworks for MAP}

The design of architectures and frameworks constitutes another active research field in MAP. Over the last years, some relevant works in MAP frameworks have been published. The work in \cite{Wilkins98} presents a complete MAP architecture for large-scale problem solving, which organizes agents into planning cells committed to a particular planning process. The TAEMS domain-independent coordination framework \cite{Lesser04} provides agents with planning capabilities, and applies the GPGP approach to coordinate them.

Other MAP architectures are based on general-purpose MAS platforms, rather than being designed from the ground up. MAS platforms, such as Magentix2 \cite{Fogues10,Argente11}  or JADE \cite{Bellifemine01}, provide the sets of services, conventions and knowledge required by agents to interact with each other. For instance, the domain-independent multi-agent system infrastructure RETSINA \cite{Sycara98} introduced a planning component \cite{Paolucci00}. Once integrated into the agents' internal architecture, this component provides them with planning capabilities.

Similarly, our MAP approach builds upon the Magentix2 MAS platform, which provides the communication services required by the agents. From this base, we introduce the additional components to provide the agents with planning capabilities and allow them to tackle MAP tasks.

\subsection{Contributions of our model}

Our novel approach to cooperative MAP can be classified into the research trend that integrates planning and coordination. The MAP system achieves two main objectives: 1) it solves complex strongly-related problems as well as loosely-coupled problems without losing generality; and 2) it allows heterogeneous agents to work under incomplete information, sharing only the critical information that affects other agents and being partially unaware of the other agents' information on the MAP task.

Our MAP approach focuses on a novel method that combines single-agent planning technologies and a refinement-based methodology. More precisely, we combine a distributed refinement planning procedure \cite{Kambhampati97} and an individual Partial-Order Planning (POP) \cite{Nguyen01,Younes03}. Agents incrementally build local refinements to a certain base plan through their local POPs, and coordinate these partial solutions through the refinement planning process. Empirical evaluation proves this method to perform effectively for both strongly-related and loosely-coupled problems.

Another key feature of our method is the ability to work under incomplete information. Unlike many MAP proposals, agents in our approach do not require to build complete knowledge bases, but they can be partially unaware of the information on the initial state and the knowledge and abilities of the rest of agents. Our \emph{PDDL3.1}-based MAP specification language \cite{Kovacs11} defines this partial visibility of the agents, allowing to specify which information can be shared with other agents for cooperation purposes. Agents exchange the shareable information with other agents through the construction of a distributed Relaxed Planning Graph \cite{Feng07} and perform planning while being partially unaware of the other agents' knowledge. This way, our proposal stresses the importance of privacy in a MAP context, as agents share only the essential information that affects other agents and are partially unaware of the information held by the rest of planning entities.

\section{Motivating example}
\label{M_example}

\begin{figure}
\centering
\includegraphics[width=10cm]{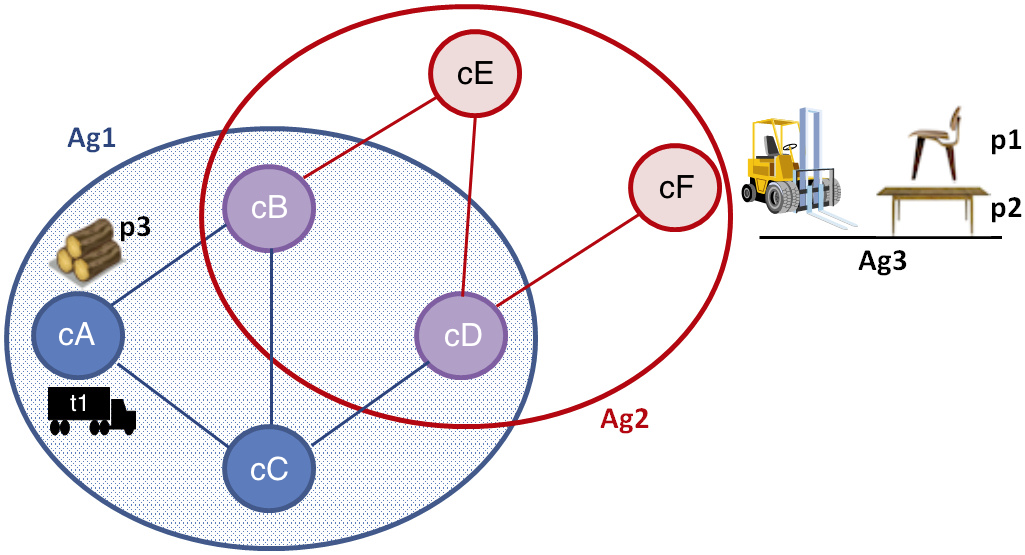}
\caption{Transportation and storage scenario}
\label{TransportTask}
\end{figure}

This section introduces the example MAP scenario we use in the following to illustrate the concepts presented throughout this paper. The example of application, depicted in Figure \ref{TransportTask}, describes a transportation and storage scenario in which two agents (Ag1 and Ag2) take the role of transport agencies and a third agent (Ag3) manages a storage facility. Transport agents deliver packages through a network of cities. In turn, the warehouse agent is in charge of storing and delivering packages to the trucks. Packages can be either raw materials or final products. Agents in the MAP task are entrusted with two different goals: deliver the final product {\ttfamily p1} to city {\ttfamily cA} and the raw material {\ttfamily p3} to city {\ttfamily cE}.

This scenario includes bidirectional links among cities that allow transport agents to move trucks from one city to another. Transport agents Ag1 and Ag2 can perform three different actions: they can load and unload packages in the trucks and they can move the trucks between cities in their working areas. Ag1 and Ag2 can only move trucks within the cities included in their working areas, depicted in Figure \ref{TransportTask} as two different circles. This way, transport agents have to interact and cooperate in order to deliver packages to a different working area.

A possible plan to solve the scenario depicted in Figure \ref{TransportTask} involves Ag1 loading the raw material {\ttfamily p3} in the truck {\ttfamily t1}. Then Ag1 would handle {\ttfamily t1} to Ag2 in {\ttfamily cB} or {\ttfamily cD}, both included in the working areas of Ag1 and Ag2, and Ag2 would take care of transporting the product to {\ttfamily cE}. This leads to a key aspect of our model: in order to promote cooperation, Ag1 should share with Ag2 the information on the position of {\ttfamily t1} once it reaches {\ttfamily cB} or {\ttfamily cD}. As we will discuss in the following section, agents will share the information that is relevant for other agents in order to successfully cooperate.

The warehouse agent Ag3 is in charge of interacting with the trucks to store raw materials and deliver final products. The warehouse has a table in which packages can be stacked and unstacked. Packages are swapped in the city in which the warehouse is placed, the exchange city. As seen in Figure \ref{TransportTask}, {\ttfamily cF} is the exchange city used by Ag2 and Ag3 to swap packages.

Ag2 and Ag3 will also share information on the packages they leave in the exchange city, which will be necessary for them to interact. For example, to accomplish the first goal of the task (transporting the final product {\ttfamily p1} to {\ttfamily cA}), Ag3 will deliver {\ttfamily p1} to the exchange city {\ttfamily cF}, informing Ag2 about the position of the package. Then, Ag2 will load {\ttfamily p1} in the truck {\ttfamily t1} and will drive {\ttfamily t1} to {\ttfamily cB} or {\ttfamily cD}. Finally, Ag1 will perform the final transportation, delivering p1 to city {\ttfamily cA}.

\section{Multi-Agent Planning architecture}
\label{Arch}

The architecture of our MAP system is depicted in Figure \ref{Architecture}. The MAP architecture basically consists of a set of agents endowed with planning capabilities and an underlying communication infrastructure that allows them to interact with each other.

All the agents share the same internal structure, and the internal planning algorithm followed by each agent is a POP procedure, so they all develop the same rationale. However, since agents handle different information and knowledge, that is, \emph{incomplete information} on the MAP task and different planning abilities, our MAP system features heterogeneous agents. In the example of application presented in section \ref{M_example}, two agents play the role of transport agencies and a third agent manages a storage facility. The first two agents will likely perform similar actions like driving vehicles from one location to another, which will be different from the planning abilities of the third agent devoted to stack and arrange packages in a warehouse. Additionally, agents will have a different view of the planning task accordingly to their abilities and initial knowledge; thus, the first two agents will have information about the trucks and roads connecting the different locations, and the third agent will manage the information about the packages and the hoists in the warehouse.

Together with the planning agents, the MAP architecture provides a set of components that allow the user to interact with the platform. The main components of the MAP architecture are:

\begin{figure}
\centering\includegraphics[width=7cm]{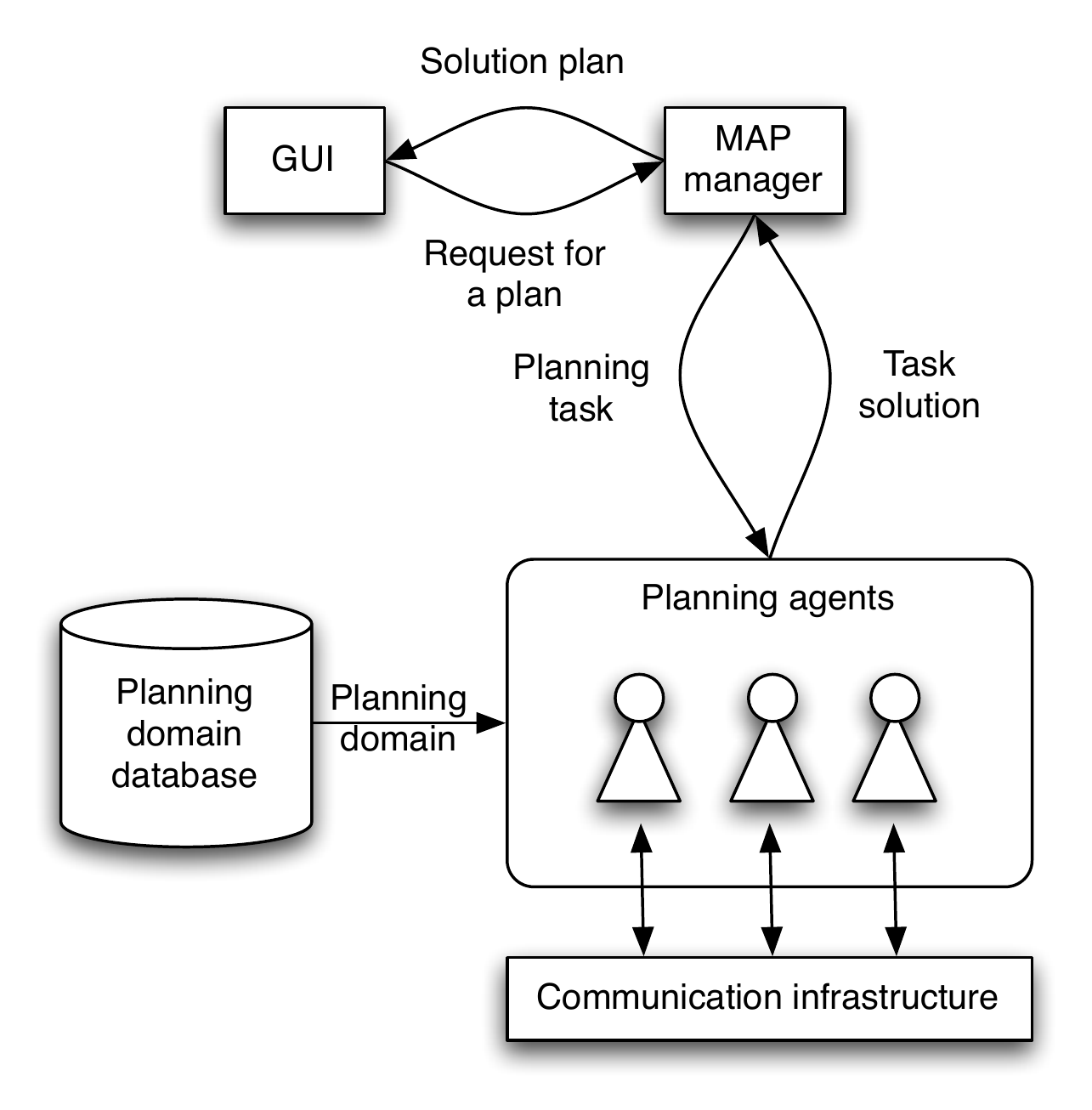}
\caption{MAP system architecture}
\label{Architecture}
\end{figure}

\begin{figure}
\centering\includegraphics[width=8cm]{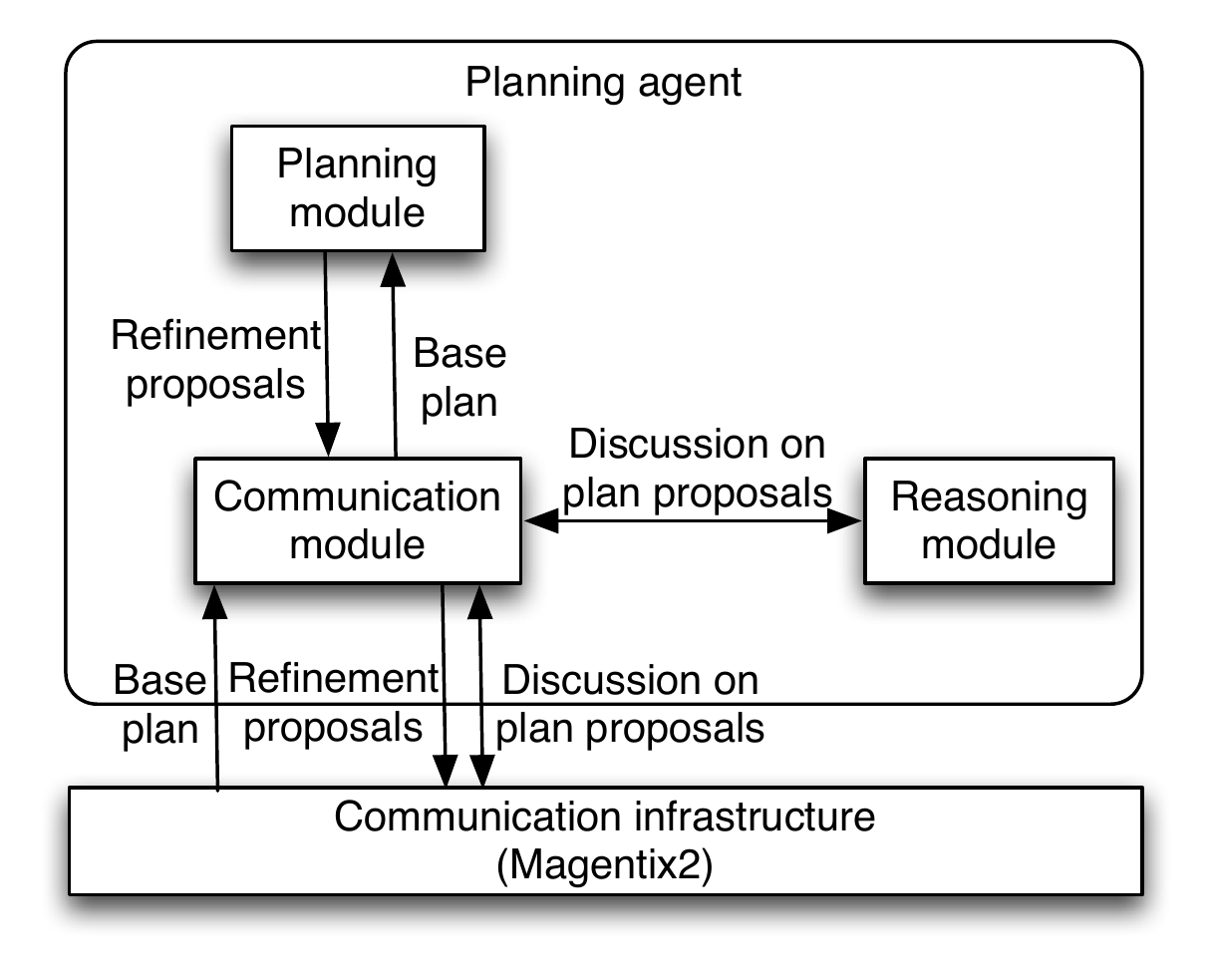}
\caption{Internal structure of a planning agent}
\label{PlanningAgent}
\end{figure}

\begin{itemize}
	\item Graphical User Interface (GUI): This component allows the user to interact with the MAP system. The user requests the resolution of a MAP task by providing, for each agent involved in the task, two input files encoded through our MAP specification language, the \emph{domain} and \emph{problem} file (see section \ref{Specification}). The first file defines the typology and the planning capabilities of the agent, while the second file defines the concrete aspects of the task it has to solve. Once a solution is found, it is displayed to the user through the GUI.
	\item MAP manager: This component interacts with the GUI by collecting the user's request for a plan and assigning the MAP task to a subset of agents that are available, i.e., they are not solving any particular planning task at the moment. Agents are fully reconfigurable and can be reused when they become available again by assigning a new MAP task to them. 	 \item Pool of planning agents: The architecture includes a pool of planning agents which all share the same internal structure shown in Figure \ref{PlanningAgent}. Agents are configurable through the domain and problem files provided by the user, which define the agents' knowledge and abilities. Once a subset of the agents in the pool receive a planning task, they start working together to find a solution plan.
	\item Communication infrastructure: Agents interact with each other through a communication infrastructure, which allows them to exchange messages by following the FIPA communication protocols \cite{Kone00}. The developed MAP system uses the Magentix2 MAS platform \cite{Fogues10} as its communication infrastructure.
\end{itemize}

The internal structure of the planning agents includes several modules to accomplish the requirements of our refinement planning approach. Through these modules, agents make plan refinements over a base plan, select the best alternative from a set of refinement plans and communicate with each other (see Figure \ref{PlanningAgent}). Although agents have the same internal structure, they have different planning abilities and visibility over the MAP task as defined in the domain and problem file provided by the user. The internal modules of a planning agent are:

\begin{itemize}
	\item Communication module: Through this module, each planning agent interacts with the rest of agents via the communication infrastructure. The communication module receives messages from the rest of agents and transmits the received information to the rest of internal modules of the planning agent. When the agent wants to communicate with other agents, this module is in charge of sending the messages through the communication infrastructure (the Magentix2 MAS platform). Hence, this module acts as an interface between the planning agent and the rest of agents in the MAP task.
	\item Planning module: This module is in charge of performing the actual planning search. It includes an embedded Partial-Order Planner which has been modified to be able to start the planning process from an incomplete plan and return valid refinements instead of complete solution plans. The planning module receives the current base plan from the communication module and returns a set of valid refinements over the base plan.
	\item Reasoning module: Agents coordination consists in evaluating the refinement plans and choosing the most promising one as the next base plan (see section \ref{Planning_reasoning}). The reasoning module of each agent receives the refinement proposals of the agents and evaluates them according to the view of the MAP task of the respective agent. Hence, this module provides agents with facilities to perform the coordination process, allowing agents to reason about the different proposals and vote for the next base plan.
\end{itemize}

In conclusion, the internal design of planning agents provides them with the basic capabilities required to solve MAP tasks. Agents use their internal components to interact with each other through the communication infrastructure, reason about plans and proceed with the next plan refinement.
\vspace{-0.1cm}
\section{Planning model}
\label{Planning model}

This section presents the MAP model upon which our planning architecture is based. It also describes the procedure followed by the agents for building and exchanging plans among them.

The following subsections describe and formalize the main components of a MAP task and outline the  Partial-Order Planning concepts used in the MAP algorithm (see section \ref{Planning_reasoning}). In order to illustrate the formal definitions introduced in this section, we provide simple examples based on the transportation MAP task presented in section \ref{M_example}. Also, for the sake of clarification of some definitions, we point out the reader to the figures of plans showed in section \ref{Problem_solving}.
\vspace{-0.1cm}
\subsection{Formalization of a MAP task}

\begin{definition}(\textbf{MAP task})
A \textbf{MAP task} is a tuple $\T = \langle\AG,$$\Obj$,$\Prop,$$\A, \I, \G,$$\rangle$. $\AG=\{1, \ldots, n\}$ is a finite non-empty set of planning agents. $\Obj$ is a finite set of objects that model the elements of the planning domain over which the planning actions can act. $\Prop$ is a finite set of state variables that model the states of the world. Each state variable $v \in \Prop$ is mapped to a finite domain of mutually exclusive values $\D_v$. Each value in a state variables's domain corresponds to an object of the planning domain, i.e. $\forall v \in \Prop$, $\D_v \subseteq \Obj$. When a value is assigned to a state variable, the pair variable-value acts as a ground atom in propositional planning. $\A$ is the set of deterministic actions of the agents. $\I$ is the set of values assigned to the state variables in $\Prop$ and represents the initial state of the MAP task $\T$. $\G$ is the set of goals of the MAP task that agents have to achieve; $\G$ represents the values that the state variables are expected to take in the final state.
\end{definition}

Information that agents have on the states of the world (problem states) is modeled through a set of ground atoms or fluents. This includes the initial state, $\I$, and the goal state, $\G$. As opposite to STRIPS-like models \cite{Fikes71}, which apply negation by failure (only positive fluents are represented, the absence of a fluent implies its negation), we allow to explicitly represent both true and false information. Thus, our model adopts the open world assumption, considering that the information which is not explicitly stored in the internal model of agents is unknown to them. Again, this also refers to the information in the initial state, $\I$, and the goals, $\G$.

\begin{definition}(\textbf{Fluent})
\label{fluent}
A ground atom or \textbf{fluent} of the problem is a tuple of the form $\langle v, d\rangle$, where $v \in \Prop$ and $d \in \D_v$. A negative fluent is of the form $\langle v, \neg d\rangle$. A positive fluent $\langle v, d\rangle$ indicates that the variable $v$ takes the value $d$, while a negative fluent $\langle v, \neg d\rangle$ indicates that the variable $v$ does not take the value $d$.
\end{definition}

As stated in Definition \ref{fluent}, a fluent relates a variable with one of the values in its domain. For instance, let ({\ttfamily at t1}) be a variable that refers to the position of a truck object {\ttfamily t1} in the example introduced in section \ref{M_example}. Possible values for this variable are the cities {\ttfamily cA}, {\ttfamily cB}, {\ttfamily cC}, {\ttfamily cD}, {\ttfamily cE} and {\ttfamily cF}. Then, a positive fluent $\langle${\ttfamily at t1}, {\ttfamily cA}$\rangle$ indicates that {\ttfamily t1} is in {\ttfamily cA} while a negative fluent $\langle${\ttfamily at t1}, $\neg${\ttfamily cA}$\rangle$ indicates that {\ttfamily t1} is not in {\ttfamily cA}.

\vspace{0,3 cm}

In our model, agents are heterogeneous as they may have different knowledge and planning capabilities. In addition, they may have \emph{incomplete information} on the MAP task as this can be distributed across agents. In this case, agents must cooperate with each other to solve the MAP task. Even though information is distributed across agents, there must be a subset of state variables shareable between agents in order to exchange the values of such variables and successfully communicate between each other. To denote the actions, goals, etc. of an agent $i \in \AG$ we will use the superscript notation $x^i$ for any such aspect $x$.

From the set of variables, $\Prop$, of the MAP task, we distinguish $\Prop^i$ as the set of variables managed by agent $i$, which includes the private variables, only known to agent $i$, and the public variables, shared with other agents. Thus, $\Prop=\{\Prop^i\}_{i=1}^n$. $D^{i}_v \subseteq D_v$ is the set of values of a variable $v \in \Prop^i$  that are visible to agent $i$. The information of the initial state of the MAP task, $\I$, is modeled through a set of positive and negative fluents. This information is distributed among agents under the assumption that agents' partial knowledge about $\I$ is consistent, i.e. there is not contradictory information among agents. Hence, $\I$ can be defined as $\I = \bigcup_{\forall i \in \AG} \I^i$. It is possible to define MAP tasks in which all the agents have a complete view of the initial state $\I$, i.e. $\forall i \in \AG,\;\I^i = \I$.

For example, Ag1 and Ag2 are two transport agents in the MAP scenario of section \ref{M_example}. Initially, Ag1 knows that the truck {\ttfamily t1} is in city {\ttfamily cA} so the fluent $\langle${\ttfamily at t1}, {\ttfamily cA}$\rangle$ is part of $\I^{Ag1}$. On the contrary, Ag2 does not know where {\ttfamily t1} is initially located, but it knows that the truck is not in city {\ttfamily cB}. Hence, the fluent $\langle${\ttfamily at t1}, $\neg${\ttfamily cB}$\rangle$ belongs to $\I^{Ag2}$, the initial state of Ag2.

\vspace{0,3 cm}

Each agent $i \in \AG$ is associated with a set, $\A^i$, of possible actions such that the set of actions of a planning task is defined as $\A = \bigcup_{\forall i \in \AG} \A^i$. An action $\alpha$ is said to be public if it is shared by two or more agents, i.e. $\alpha \in A^i \wedge \alpha \in A^j$, $i \neq j$. $\alpha \in \A^i$ is private to agent $i$ iff $\alpha \not\in \A^j, \forall j \neq i$. An action $\alpha \in \A^i$ denotes that agent $i$ has the capability expressed in $\alpha$. If $\alpha$ forms part of the final plan then agent $i$ is also responsible of executing $\alpha$.

\begin{definition}(\textbf{Planning rule or action})
A \textbf{planning rule or action} $\alpha \in \A$ is a tuple $\langle PRE(\alpha)$, $EFF(\alpha) \rangle$. $PRE(\alpha)=\{p_1, \ldots, p_n\}$ is a set of fluents that represents the preconditions of $\alpha$, while $EFF(\alpha)=\{e_1, \ldots, e_m\}$ is a set of operations of the form $(v = d)$ or $(v \not= d)$, $v \in \Prop$, $d \in \D_v$, that represent the consequences of executing $\alpha$.
\end{definition}

An action $\alpha$ may belong to different agents, i.e. $\alpha \in \A^i$ and $\alpha \in \A^j$, $i \neq j$. Executing an action $\alpha$ in a world state $S$ gives rise to a new world state $S'$ generated as the result of applying $EFF(\alpha)$ over $S$. Particularly:

\begin{itemize}
\item An operation $(v = d) \in EFF(\alpha)$ implies the addition of a fluent $\langle v, d\rangle$ and a set of fluents $\langle v, \neg d'\rangle$, $\forall d' \in \D_v\;|\;d' \not= d$, to the world state $S'$. If $\langle v, d'\rangle \in S$ or $\langle v, \neg d\rangle \in S$, $d' \not= d$, the operation $(v = d)$ also implies that the fluents $\langle v, d'\rangle$ or $\langle v, \neg d\rangle$ will not be present in $S'$. For example, suppose that agent Ag1 knows that the truck {\ttfamily t1} can be placed at the cities {\ttfamily cA}, {\ttfamily cB}, {\ttfamily cC} and {\ttfamily cD}, i.e., $\D^{Ag1}_{at\,t1} = \{${\ttfamily cA}, {\ttfamily cB}, {\ttfamily cC}, {\ttfamily cD}$\}$. If Ag1 knows a positive fluent $\langle${\ttfamily at t1}, {\ttfamily cA}$\rangle$, it also knows the negative fluents $\langle${\ttfamily at t1}, $\neg${\ttfamily cB}$\rangle$, $\langle${\ttfamily at t1}, $\neg${\ttfamily cC}$\rangle$ and $\langle${\ttfamily at t1}, $\neg${\ttfamily cD}$\rangle$.

\item An operation $(v \not= d) \in EFF(\alpha)$ implies the addition of a fluent $\langle v, \neg d\rangle$ to the world state $S'$. If $\langle v, d\rangle \in S$, the operation $(v \not= d)$ also entails that the fluent $\langle v, d\rangle$ will not be present in $S'$. Note that the only existence of a fluent $\langle v, \neg d\rangle$ in a state indicates that the value of the variable $v$ is not known in such a state and, consequently, the rest of values in $\D_v$, except for $d$, are unknown values. For example, if the fluent $\langle${\ttfamily at t1}, $\neg${\ttfamily cB}$\rangle$ is in the world state of agent Ag2, then the agent only knows that the truck {\ttfamily t1} is not in city {\ttfamily cB} but the agent is not aware of the actual position of the truck. Thus, whether {\ttfamily t1} is in {\ttfamily cA}, {\ttfamily cC} or {\ttfamily cD} is unknown to Ag2.
\end{itemize}

The set of preconditions of an action $\alpha$, $PRE(\alpha)$, defines the fluents that must hold in a world state $S$ for that $\alpha$ is applicable in this state. A positive precondition of the form $\langle v, d\rangle$ indicates that the fluent $\langle v, d\rangle$ must hold in $S$, while a negative precondition $\langle v, \neg d\rangle$ indicates that the fluent $\langle v, \neg d\rangle$ must hold in $S$. Note that the existence of a positive fluent $\langle v, d\rangle$ also implies the existence of a negative fluent $\langle v, \neg d'\rangle$ for the rest of values in the variable's domain, i.e. $(\exists \langle v, d\rangle \in S) \Rightarrow (\forall d' \in \D_v$, $d' \not= d$, $\exists \langle v, \neg d'\rangle \in S)$.

Additionally, agents use a utility function $\F$ to evaluate the quality of the plan proposals. For each agent $i$, $\F$ assigns a cost, $\cost(view^i(\Pi)) \in \mathbb{R}_{0}^{+}$, to each plan proposal $\Pi$ according to the view that agent $i$ has of that plan, $view^i(\Pi)$. Finally, the private goals of an agent $i$, $\PG^i$, are fluents that agent $i$ is interested in attaining. Private goals are encoded as soft constraints \cite{Gerevini06a}, as it is not mandatory that agents achieve them.

\subsection{Concepts on Partial-Order Planning}

Our MAP model can be regarded as a multi-agent refinement planning framework, a general method based on the refinement of the set of all possible plans \cite{Kambhampati97}. An agent proposes a plan $\Pi$ that typically enforces some of the goals that have not yet been supported (see definition \ref{og}); then, the rest of agents collaborate on the refinement of $\Pi$ by solving some of these pending goals in $\Pi$. This way, agents cooperatively solve the MAP task by consecutively refining an initially empty plan.

In this context, Partial-Order Planning (POP) \cite{Barrett94} arises as a suitable approach to address refinement planning, since it is focused on solving the pending goals progressively. Consequently, agents in our MAP approach plan concurrent actions through the adoption of the POP paradigm. In the following, we provide some basic definitions concerning single-agent POP and its adaptation to a MAP context.

\subsubsection{Single-agent Partial-Order Planning}

\begin{definition} (\textbf{Partial plan})
A \textbf{partial plan} is a tuple $\Pi = \langle \Delta, \Or, \CL$$\rangle$. $\Delta \subseteq \A$ is the set of actions in $\Pi$. $\Or$ is a set of ordering constraints ($\prec$) on $\Delta$. $\CL$ is a set of causal links over $\Delta$. A causal link is of the form $\alpha \stackrel{\langle v,d\rangle}{\rightarrow} \beta$ or $\alpha \stackrel{\langle v,\neg d\rangle}{\rightarrow} \beta$, where $\alpha \in \A$ and $\beta \in \A$ are actions in $\Delta$. $\alpha \stackrel{\langle v,d\rangle}{\rightarrow} \beta$ indicates that there is an operation $(v = d)$ such that $v \in \Prop$, $d \in \D_v$, $(v = d) \in EFF(\alpha)$ and a fluent $\langle v, d\rangle \in PRE(\beta)$. $\alpha \stackrel{\langle v,\neg d\rangle}{\rightarrow} \beta$ indicates that there is a fluent $\langle v, \neg d\rangle$ such that $v \in \Prop$, $d \in \D_v$, $\langle v, \neg d\rangle \in PRE(\beta)$ supported by an operation $(v \not= d) \in EFF(\alpha)$ or an operation $(v = d') \in EFF(\alpha)$, $d' \in \D_v$, $d' \not= d$.
\end{definition}

This definition of partial plan represents the mapping of a plan into a directed acyclic graph, where $\Delta$ represents the nodes of the graph (actions) and $\Or$ and $\CL$ are sets of directed edges representing the precedences and causal links among these actions, respectively.

An \emph{empty} partial plan is defined as $\Pi_0 = \langle \Delta_0$, $\Or_0$, $\CL_0 \rangle$, where $\Delta_0$ contains $\alpha_0$ and $\alpha_f$, the initial and final action of the plan, respectively. $\alpha_0$ and $\alpha_f$ are fictitious actions that do not belong to the action set of any particular agent. $\Or_0$ contains the constraint $\alpha_0 \prec \alpha_f$ and $\CL_0$ is an empty set. This way, a plan $\Pi$ for any given MAP task $\T$ will always contain the two fictitious actions such that $PRE(\alpha_0)=\emptyset$ and $EFF(\alpha_0)=\I$, $PRE(\alpha_f)=\G$, and $EFF(\alpha_f)=\emptyset$; i.e. $\alpha_0$ represents the initial situation of the MAP task $\T$, and $\alpha_f$ represents the global goals of $\T$.

Assuming that $\G \neq \emptyset$, an empty plan is said to be incomplete if the preconditions of $\alpha_f$ are not yet supported through a causal link. The POP process is aimed at introducing causal links to support these preconditions, also called \emph{open goals}.

\begin{definition} (\textbf{Open goal})
\label{og}
An \textbf{open goal} in a partial plan $\Pi = \langle \Delta$, $\Or$, $\CL \rangle$ is a fluent $og$ of the form $\langle v, d\rangle$ or $\langle v, \neg d\rangle$, such that $v \in \Prop$, $d \in \D_v$, $og \in PRE(\beta)$, $\beta \in \Delta$, and $\nexists \alpha \in \Delta / \alpha \stackrel{og}{\rightarrow} \beta \in \CL$, i.e., an open goal $og$ is a precondition of ab action $\beta$ in the plan $\Pi$ that is not yet supported by a causal link $\alpha \stackrel{og}{\rightarrow} \beta \in \CL$. $\openGoals(\Pi)$ denotes the set of open goals in $\Pi$. A plan is \textbf{incomplete} if it has open goals. Otherwise, we say it is a \textbf{complete} plan.
\end{definition}

As the POP search progresses, the causal links in a partial plan may become unsafe as a result of the introduction of a new action which is not ordered with respect to the causal link. These conflicts are called \emph{threats}.

\begin{definition} (\textbf{Threat})
A \textbf{threat} in a partial plan $\Pi = \langle \Delta$, $\Or$, $\CL \rangle$ represents a conflict between an action of the plan and a causal link. An action $\gamma$ causes a threat over a causal link $\alpha \stackrel{\langle v,d\rangle}{\rightarrow}  \beta$ if $((v = d') \in EFF(\gamma) \vee (v \not= d) \in EFF(\gamma))$, where $v \in \Prop$, $d \in \D_v$, $d' \in \D_v$ and $d \not= d'$, and there is neither an ordering constraint $\gamma \prec \alpha$ nor $\beta \prec \gamma$. The action $\gamma$ will cause a threat over a causal link of the form $\alpha \stackrel{\langle v,\neg d\rangle}{\rightarrow} \beta$ if $(v = d) \in EFF(\gamma)$, where $v \in \Prop$, $d \in \D_v$, and there is neither an ordering $\gamma \prec \alpha$ nor $\beta \prec \gamma$. $Threats(\Pi)$ denotes the set of threats in $\Pi$.
\end{definition}

A threat $t \in Threats(\Pi)$ can be solved by \emph{promoting} or \emph{demoting} the threatening action $\gamma$ with respect to the causal link $\alpha \stackrel{\langle v,d\rangle}{\rightarrow} \beta$ or $\alpha \stackrel{\langle v,\neg d\rangle}{\rightarrow} \beta$, i.e. introducing an ordering constraint $\gamma \prec \alpha$ or $\beta \prec \gamma$. Threats and open goals are referred to as the \emph{flaws} of a partial-order plan. The POP process is guided by solving the pending flaws of an initially empty partial plan.

Figure \ref{Refinement00} in section \ref{Problem_solving} depicts a refinement plan for the example introduced in section \ref{M_example}. This refinement plan includes a causal link {\ttfamily Init} $\stackrel{\langle \mathsf{at\,t1,\,cA} \rangle}{\rightarrow}$ {\ttfamily load t1 p3 cA}. Suppose that a new action {\ttfamily drive t1 cA cB}, that causes the truck {\ttfamily t1} to move from {\ttfamily cA} to {\ttfamily cB}, is added to the refinement plan and that this new action is not ordered with respect to {\ttfamily load t1 p3 cA}. In this case, $(${\ttfamily at t1} $=$ {\ttfamily cB}$)\,\in EFF(${\ttfamily drive t1 cA cB}$)$. This effect causes a threat over the previous causal link, as it introduces a fluent $\langle${\ttfamily at t1, cB}$\rangle$ that affects the value of the variable ({\ttfamily at t1}). This threat can be solved by introducing an ordering constraint {\ttfamily load t1 p3 cA} $\prec$ {\ttfamily drive t1 cA cB}, i.e., demoting the threatening action {\ttfamily drive t1 cA cB} with respect to the causal link.

\subsubsection{Multi-agent Partial-Order Planning}
\label{MAPOP}

Agents in our MAP model cooperate on the refinement of a base plan $\Pi$ by proposing \emph{refinement steps} that solve some open goals in $\Pi$. This way, agents cooperatively solve the MAP task by consecutively refining $\Pi$, the initially empty base plan.

\begin{definition} (\textbf{Refinement step})
A refinement step $\Pi^i$ devised by an agent $i$ over a base plan $\Pi_{g}$, where $g \in \openGoals(\Pi_{g})$, is a triple $\Pi^i = \langle \Delta^i, OR^i, CL^i \rangle$, where $\Delta^i \subseteq \A$ is a set of actions and $OR^i$ and $CL^i$ are sets of orderings and causal links over $\Delta^i$, respectively. $\Pi^i$ is a threat-free partial plan that solves $g$ as well as all the new open goals of the form $\langle v, d\rangle$ or $\langle v, \neg d\rangle$ that arise from this resolution and can only be achieved by agent $i$, where $(v \in \Prop^i) \wedge (v \not\in \Prop^j, \forall j \neq i)$. That is, when solving a goal of a base plan, agents only accomplish the new open goals concerning their private fluents, leaving public goals unresolved. In other words, the refinement method only iterates over the public fluents. Let $g \in \openGoals(\Pi_{g})$ be a fluent of the form $\langle v, d\rangle$ or $\langle v, \neg d\rangle$; an agent $i$ proposes a refinement step over $\Pi_{g}$ iff $v \in \Prop^i$.
\end{definition}

In our MAP approach partial plans are multi-agent concurrent plans as two or more actions can be concurrently executed by different agents. Some MAP models adopt a simple form of concurrency: two concurrent actions are mutually consistent if none of them changes the value of a state variable that the other relies on or affects, too \cite{Brenner09}. We impose the additional concurrency constraint that the preconditions of two actions have to be consistent \cite{Boutilier01} for these two actions to be mutually consistent. This definition of concurrency is straightforwardly extended to a joint action $\alpha=\langle \alpha_1, \ldots, \alpha_n\rangle$.

\begin{definition} (\textbf{Mutually consistent actions})
\label{Consistent}
Two concurrent actions $\alpha \in \A^i$ and $\beta \in \A^j$ are \textbf{mutually consistent} if none of the following conditions holds:

\begin{itemize}
\item $\;\exists (v = d) \in EFF(\alpha)$ and $\exists (\langle v, d'\rangle \in PRE(\beta) \vee \langle v, \neg d\rangle \in PRE(\beta))$, where $v \in \Prop^i \cap \Prop^j$, $d \in \D^{i}_v \cap \D^{j}_v$, $d' \in \D^{j}_v$ and $d \not= d'$, or vice versa.
\item $\;\exists (v = d) \in EFF(\alpha)$ and $\exists ((v = d') \in EFF(\beta) \vee (v \not= d) \in EFF(\beta))$, where $v \in \Prop^i \cap \Prop^j$, $d \in \D^{i}_v \cap \D^{j}_v$, $d' \in \D^{j}_v$ and $d \not= d'$, or vice versa.
\item $\;\exists \langle v, d\rangle \in PRE(\alpha)$ and $\exists (\langle v, d'\rangle \in PRE(\beta) \vee \langle v, \neg d\rangle \in PRE(\beta))$, where $v \in \Prop^i \cap \Prop^j$, $d \in \D^{i}_v \cap \D^{j}_v$, $d' \in \D^{j}_v$ and $d \not= d'$, or vice versa.
\end{itemize}
\end{definition}

Going back to our example in section \ref{M_example}, two concurrent actions {\ttfamily drive t1 cA cB}, planned by agent Ag1, and {\ttfamily drive t1 cA cC}, planned by agent Ag2, are mutually inconsistent as $(${\ttfamily at t1} $=$ {\ttfamily cB}$)\,\in EFF(${\ttfamily drive t1 cA cB}$)$ and $\langle${\ttfamily at t1, cC}$\rangle\,\in PRE(${\ttfamily drive t1 cC cB}$)$ (the first condition in Definition \ref{Consistent} holds). Concurrent actions {\ttfamily drive t1 cA cB} and {\ttfamily drive t1 cA cC} are also mutually inconsistent as $(${\ttfamily at t1} $=$ {\ttfamily cB}$)\,\in EFF(${\ttfamily drive t1 cA cB}$)$ and $(${\ttfamily at t1} $=$ {\ttfamily cC}$)\,\in EFF(${\ttfamily drive t1 cA cC}$)$ (second condition holds). Finally, concurrent actions {\ttfamily drive t1 cA cB} and {\ttfamily drive t1 cC cB} are mutually inconsistent as $\langle${\ttfamily at t1, cA}$\rangle\,\in PRE(${\ttfamily drive t1 cA cB}$)$ and $\langle${\ttfamily at t1, cC}$\rangle\,\in PRE(${\ttfamily drive t1 cC cB}$)$ (third condition holds).

\vspace{0,3 cm}

As agents address concurrency inconsistencies through the detection of threats over the causal links of their plans, concurrency is ensured among private actions since a refinement step put forward by an agent is always a threat-free plan. However, concurrency issues between two public actions introduced by different agents do not arise until their preconditions are fully supported through causal links. This way, it is not possible to ensure that two concurrent actions are mutually consistent until their preconditions are fully supported. Thus, our notion of multi-agent concurrent plan distinguishes between public and private actions when dealing with concurrency.

\begin{definition} (\textbf{Multi-agent concurrent plan})
A partial plan $\Pi = \langle \Delta, \Or,$ $\CL \rangle$ is a \textbf{multi-agent concurrent plan} if for every pair of unequal, concurrent, public actions $\alpha$ and  $\beta$, $\alpha \not= \beta$, $\forall p_\alpha \in PRE(\alpha), p_\alpha \not \in \openGoals(\Pi)$, $\forall p_\beta \in PRE(\beta), p_\beta \not \in \openGoals(\Pi)$, $\alpha$ and $\beta$ are mutually consistent.
\end{definition}

\begin{definition} (\textbf{Refinement plan})
A refinement plan $\Pi$ devised by an agent $i$ over a base plan $\Pi_{g}$ is a concurrent multi-agent plan that results from the composition of $\Pi_{g}$ and a refinement step $\Pi^i$ proposed by agent $i$. $\Pi$ is defined as $\Pi=\Pi_{g} \circ \Pi^i$, where $\circ$ represents the composition operation.
\end{definition}

Thus, an agent $i$ can build a refinement plan $\Pi$ upon a base plan $\Pi_{g}$ by composing $\Pi_{g}$ and a refinement step $\Pi^i$ that solves at least $g \in \openGoals(\Pi_{g})$, i.e. $\Pi=\Pi_{g} \circ \Pi^i$. As previously mentioned, refinement steps are always threat-free and their actions are mutually consistent. Hence, if a refinement step brings about a concurrency inconsistency or a threat on the composite plan, the proposer agent is responsible for addressing such a flaw. If an agent is not able to come up with a consistent refinement plan, then it refrains from suggesting it. In case no refinements for a base plan can be found, the base plan is said to be a dead-end plan.

\begin{definition} (\textbf{Dead-end plan})
A plan $\Pi$ is called a \textbf{dead-end plan} if $\exists g \in \openGoals(\Pi)$ and there is no refinement step $\Pi^i$ such that $g \not\in \openGoals(\Pi \circ \Pi^i)$; that is, no refinement step solves the open goal $g$.
\end{definition}

\begin{definition} (\textbf{Solution plan})
A multi-agent concurrent plan $\Pi$ is a \textbf{solution plan} for a planning task $\T$ if $\openGoals(\Pi) = \emptyset$ ($\Pi$ is a complete plan), $Threats(\Pi) = \emptyset$, and every pair of actions $\alpha, \beta \in \Pi$ are mutually consistent.
\end{definition}

That is, a solution plan is a complete multi-agent concurrent plan. Note that we require $\Pi$ to be a complete plan so it cannot have pending open goals. Consequently, the preconditions of the fictitious final action $\alpha_f$ will also hold thus guaranteeing that $\Pi$ solves the MAP task $\T$. For instance, Figure \ref{Solution} in section \ref{Problem_solving} shows a solution plan for the MAP task presented in section \ref{M_example}. The different shapes of the actions indicate which agent has proposed them. The solution plan in Figure \ref{Solution} is a complete, concurrent plan to which all the agents in the MAP task have contributed.

\section{Planning language for MAP tasks}
\label{Specification}

In our MAP system, we define the agents' planning tasks through several specification files. These files encode the information of the agent on the MAP task, namely the variables, $\Prop^i$; the objects associated to the variables, $\Obj^i$; the planning actions, $\A^i$; and the initial state of the agent, $\I^i$. All this information is written in a planning definition language.

Traditionally, planning has been regarded as a single-agent problem, where only one centralized planning entity is required. MAP presents new requirements and challenges that are not present in classical, centralized planning. Planning agents in our MAP model can withhold their private information, and decide which information to share with the rest of agents. In addition, planning agents can have private individual objectives besides the goals of the planning task. Therefore, the planning language must provide support to allow us to define shareable information and private goals.

Planning definition languages have experienced a remarkable evolution over the last years, continuously increasing their expressivity through the addition of new features. Our MAP language is based on \emph{PDDL3.1} \cite{Kovacs11}, the most recent version of \emph{PDDL} \cite{Ghallab98}, which was introduced in the context of the 2008 International Planning Competition. Unlike its predecessors, that model a planning domain through logical predicates, \emph{PDDL3.1} also incorporates state variables by adding fluents that map a tuple of objects to an object of the planning task. We have extended the \emph{PDDL3.1} language with some new structures to model the multi-agent features of a planning task.

In single-agent PDDL language, the user writes two files, one containing the \emph{domain} of the task and another one containing the data of the \emph{problem} to be solved. The domain file describes the planning actions, the types of objects and the state variables of the task, while the problem file details the current objects of the task, the initial state (the initial values of the state variables) and the task goals. These files have a similar structure to their \emph{PDDL} counterparts, and reflects the additional information required by MAP tasks.

In our MAP system, each agent has a domain and a problem file that model, respectively, the typology of the planning agent and its particular vision of the MAP task. The domain and problem files also include the information that is shared among agents. The {\ttfamily shared-data} structure allows the problem designer to define which fluents will be shared by each agent and with whom. Through this structure, the designer can define the \emph{incomplete information} of the agent. This way, the domain knowledge of the agents can be modeled (or specified) from a complete unawareness to a full visibility of the domain. Additionally, since agents in MAP may have both global and local goals, this information is modeled through the structures {\ttfamily global-goal} and {\ttfamily private-goal}. Finally, we have included an additional {\ttfamily multi-functions} structure in order to simplify the specification of fluents in the initial state of an agent.

The following subsections analyze the structures that cover the requirements of MAP domains, i.e. modeling the data shared among agents, and the definition of local and global goals. The last subsection provides an example that describes the encoding of the MAP task introduced in section \ref{M_example} with our language.

\subsection{Shared data}
\label{Shared_data}

The {\ttfamily shared-data} structure, located on the agent's problem file, determines which fluents are shareable and with which agent or agents they will be shared. As shown in section \ref{Initial}, this structure directly affects the initial information exchange that agents perform before planning, and it also defines the partial view of the planning task of each agent.

As agents only exchange fluents, in the {\ttfamily :shared-data} structure the problem designer specifies the fluents that the agent can share and with which agents. The {\ttfamily shared-data} structure has the following BNF syntax:

\renewcommand{\ttdefault}{pcr}
{\ttfamily\begin{lstlisting}
<shared-data-def>  ::= (:shared-data <share-def>+)
<share-def>        ::= (<atom-formula-def>+ [- <agent-def>?])
<agent-def>        ::= <agent> | (either <agent> <agent>+)
<agent>            ::= <name>
<atom-formula-def> ::= (<predicate> <typed-list(element)>)
<atom-formula-def> ::= (= <object-fluent> <object>)
<predicate>        ::= <name>
<object-fluent>    ::= (<name> <object>*)
<object>           ::= <name>
<element>          ::= <variable> | <constant>
<variable>         ::= ?<name>
<constant>         ::= <name>
<typed-list(x)>    ::= x*
\end{lstlisting}}
\renewcommand{\ttdefault}{cmtt}

As the BNF syntax shows, it is possible to define fluents or predicates within the {\ttfamily :shared-data} section and associate them to one, some or all the agents in the system (if {\ttfamily agent} is not specified, the predicates or fluents are shared with all the agents).

\subsection{Private and global goals}

A particularity of the MAP approach when compared to traditional planning is the fact that agents have private and global goals. To reflect this information in the model, the {\ttfamily private-goal} and {\ttfamily global-goal} structures have been included into the problem files. Similarly to the {\ttfamily goal} section in \emph{PDDL3.1}, goals can be modeled through predicates or fluents. The {\ttfamily private-goal} and {\ttfamily global-goal} structures use the following BNF syntax:
\newpage
\renewcommand{\ttdefault}{pcr}
{\ttfamily\begin{lstlisting}
<private-goal-def>  ::= (:private-goal <predicate-def>)
<global-goal-def>   ::= (:global-goal <predicate-def>)
<predicate-def>     ::= <atom-formula-def>
<predicate-def>     ::= (and <atom-form-def> <atom-form-def>+)
<predicate-def>     ::= (or <atom-form-def> <atom-form-def>+)
<atom-form-def>     ::= (<predicate> <typed-list(element)>)
<atom-form-def>     ::= (= <object-fluent-def> <object>)
<predicate>         ::= <name>
<object-fluent-def> ::= (<name> <object>*)
<object>            ::= <name>
<element>           ::= <variable> | <constant>
<variable>          ::= ?<name>
<constant>          ::= <name>
<typed-list(x)>     ::= x*
\end{lstlisting}}
\renewcommand{\ttdefault}{cmtt}

As shown in the BNF syntax description, both global and local goals are described as conjunctions or disjunctions of fluents and predicates, or rather as a single fluent or predicate.

\subsection{Encoding example}
\label{Example_modeling}

This subsection describes the encoding of the MAP task presented in section \ref{M_example} with our MAP language. This MAP task describes a transportation and storage scenario in which two agents (Ag1 and Ag2) take the role of transport agencies and an agent (Ag3) manages a storage facility. Transport agents deliver packages through a network of cities, while the warehouse agent stores and loads the packages in trucks. In the following, we provide a description of the domain and problem files of the agents for this task, stressing the specification of shareable information.

Planning agents receive two different description files, namely the domain and problem file. The domain file contains a general description of the capabilities of the agent, including the actions that the agent can perform and the predicates and functions it can manage. All agents of the same type share the same domain file, e.g. transport agents Ag1 and Ag2 in this example receive the same domain file. The problem file models the concrete problem assigned to each agent, including a description of the objects of the task, the initial situation and the global goals of the task as well as private goals of the agent. Each agent receives its particular problem file.

\subsubsection{Domain files}

The domain file for transport agents specifies bidirectional links among cities, which allow trucks to move from one city to another. Trucks can only travel within the cities included in their working areas, depicted in Figure \ref{TransportTask} with two circles. This way, transport agents have to interact and cooperate in order to deliver packages to a different area. The domain file for transport agents is modeled as follows:
\newpage
\renewcommand{\ttdefault}{pcr}
{\ttfamily\small\begin{lstlisting}
(define (domain Transport)
 (:requirements :typing :equality :fluents)
 (:types truck package agent city - object
         raw-material final-product - package)
 (:predicates (empty ?c - city))
 (:functions (at ?t - truck) - city
             (pos ?p - package) - (either city truck)
 )
 (:@\color{blue}\bfseries{multi-functions}@ (link ?c - city) - city
                 (area) - city
 )
 (:action load
  :parameters (?t - truck ?p - package ?c - city)
  :precondition (and (member (area) ?c)(= (at ?t) ?c)(= (pos ?p) ?c))
  :effect (and (assign (pos ?p) ?t)(empty ?c))
 )
 (:action unload
  :parameters (?t - truck ?p - package ?c - city)
  :precondition (and (empty ?c)(member (area) ?c)
                (= (at ?t) ?c)(= (pos ?p) ?t))
  :effect (and (assign (pos ?p) ?c)(not (empty ?c)))
 )
 (:action drive
  :parameters (?t - truck ?c1 ?c2 - city)
  :precondition (and (member (area) ?c1)(member (area) ?c2)
                     (member (link ?c1) ?c2)(= (at ?t) ?c1))
  :effect (assign (at ?t) ?c2)
 )
)
 \end{lstlisting}}
\renewcommand{\ttdefault}{cmtt}

The domain file shown above is structured similarly as a regular \emph{PDDL3.1} file. The main sections of the file are highlighted in bold. The {\ttfamily :requirements} section indicates the \emph{PDDL} features that have been used to encode the domain information. {\ttfamily :types} describes the object-type hierarchy of this particular domain. As it can be seen, the planning domain of transport agents includes four different types of objects, namely {\ttfamily truck}, {\ttfamily agent}, {\ttfamily city} and {\ttfamily package}. A {\ttfamily package} can be either a {\ttfamily raw-material} or a {\ttfamily final-product}.

Structures {\ttfamily :predicates}, {\ttfamily :functions} and {\ttfamily :multi-functions} define the state variables used in the transport domain. During the planning process, these variables will be instantiated to objects defined in the transport agents' problems, thus giving rise to the fluents that will be used throughput the planning process. For instance, let us consider the function {\ttfamily (at ?t - truck) - city}, where ({\ttfamily at ?t}) is a state variable and {\ttfamily city} is the type of its domain values. Given a {\ttfamily truck} object {\ttfamily t1} and a {\ttfamily city} object {\ttfamily c1}, the previous function will result in a fluent of the form {\ttfamily (= (at t1) c1)}, which indicates that {\ttfamily t1} is located at {\ttfamily c1}.

The domain file of transport agents include the following predicates, functions and multi-functions: {\ttfamily empty} is a predicate to indicate whether a {\ttfamily city} is empty or already contains a {\ttfamily package} (a {\ttfamily city} can only have one {\ttfamily package} simultaneously); function {\ttfamily at} returns the {\ttfamily city} in which a certain {\ttfamily truck} is placed; function {\ttfamily pos} describes the position of a {\ttfamily package}, either a {\ttfamily truck} or a {\ttfamily city}; multi-function {\ttfamily link} returns the outcoming connections (roads) from a certain {\ttfamily city}; and {\ttfamily area} describes the working area of an agent in terms of the cities it can drive a {\ttfamily truck} to.

The last portion of a \emph{PDDL3.1} domain file defines the abilities of the agent, i.e., the actions it can perform. Actions are described through its {\ttfamily parameters} (objects that take part in the action), {\ttfamily preconditions} (predicates and functions that must hold for the action to be applicable) and {\ttfamily effects} (predicates and functions that describe the consequences of applying the action). As in the case of predicates, functions and multi-functions, actions are described through state variables. In particular, preconditions encode queries on fluents that check whether a variable takes on a particular value, and effects encode assignment operations on fluents to make a state variable take on a value.

As described in the domain file, transport agents can perform three different actions: {\ttfamily load} and {\ttfamily unload} a {\ttfamily package} into/from a {\ttfamily truck}, and {\ttfamily drive} a {\ttfamily truck} from a {\ttfamily city} to another one of the {\ttfamily agent}'s {\ttfamily area}.

The domain file for warehouse agents is similar to the classical \emph{blocksworld} domain, in which packages can be stacked and unstacked on/from the table or other packages. In this case, only one pile of packages can be stacked on the table, and there are two types of packages, raw materials and final products. The transportation and storage scenario depicted in Figure \ref{TransportTask} includes two final products (packages {\ttfamily p1} and {\ttfamily p2}) and a raw material (package {\ttfamily p3}). The warehouse agent delivers final products to the city in which the warehouse is placed (the exchange city, {\ttfamily cF} in Figure \ref{TransportTask}), and acquires raw materials that are unloaded from the trucks in the exchange city. Following, we show a sketch of the warehouse domain file encoding:

\renewcommand{\ttdefault}{pcr}
{\ttfamily\small\begin{lstlisting}
(define (domain Warehouse)
 (:requirements :typing :equality :fluents)
 (:types package agent city table hoist - object
         raw-material final-product - package)
 (:predicates (empty ?c - city)
              (clear ?p - (either package table hoist))
              (exchange-city ?c - city)
 )
 (:functions (pos ?p - package) - (either city package table hoist))
 (:action acquire
  :parameters (?p - raw-material ?c - city ?h - hoist)
  :precondition (and (= (pos ?p) ?c)(clear ?h)(exchange-city ?c))
  :effect (and (assign (pos ?p) ?h)(not (clear ?h))(empty ?c))
 )

 ...

)
\end{lstlisting}}
\renewcommand{\ttdefault}{cmtt}

As the transport agents, warehouse agents manage {\ttfamily city}, {\ttfamily hoist} and {\ttfamily package} objects. Additionally, warehouse agents consider {\ttfamily table} and {\ttfamily hoist} objects. The {\ttfamily hoist} is used to {\ttfamily deliver} and {\ttfamily acquire} packages, while the {\ttfamily table} is used to {\ttfamily stack} and {\ttfamily unstack} packages within the warehouse.

Warehouse agents perform the four actions indicated above: they can {\ttfamily stack} and {\ttfamily unstack} a {\ttfamily package} on top/from a clear {\ttfamily table} or {\ttfamily package}; and can also {\ttfamily acquire} and {\ttfamily deliver} a {\ttfamily package} from/to the {\ttfamily exchange-city} by using a {\ttfamily hoist}. The sketch of the domain file illustrates the encoding of the {\ttfamily acquire} action.

\subsubsection{Problem files}
\label{pfiles}

Each agent receives its own problem file that models the particular objects managed by the agent, the initial situation known to the agent and the global and private goals that the agent must achieve. Moreover, the problem files include the definition of the shareable fluents and with which agents they can be shared.

We now explain the problem file of transport agent Ag1 (this problem will be later used to illustrate the construction of the dis-RPG). Problem files describe the initial state of the task by including both the positive and negative information known by the agent. This way, the information not represented in the problem file is unknown to the agent. Ag1's problem file is encoded as follows:

\renewcommand{\ttdefault}{pcr}
{\ttfamily\small\begin{lstlisting}
(define (problem Ag1)
(:domain Transport)
(:objects
     Ag1 Ag2 Ag3 - agent
     t1 - truck
     cA cB cC cD cE cF - city
     p3 - raw-material
     p1 p2 - final-product)
(:@\color{blue}\bfseries{shared-data}@
     (empty ?c - city) - (either Ag2 Ag3)
     ((at ?t - truck) - city)
     ((pos ?p - package) - (either city truck)) - Ag2
     ((pos ?p - package) - city) - Ag3
)
(:init
     (empty cB) (empty cC) (empty cD) (not (empty cA))
     (= (at t1) cA) (not (= (at t1) cB)) (not (= (at t1) cC))
     (not (= (at t1) cD)) (= (pos p3) cA) (not (= (pos p3) cB))
     (not (= (pos p3) cC)) (not (= (pos p3) cD))
     (= (link cA) {cB cC}) (not (= (link cA) {cA cD}))
     (= (link cB) {cA cC}) (not (= (link cB) {cB cD}))
     (= (link cC) {cA cB cD}) (not (= (link cC) {cC}))
     (= (link cD) {cC}) (not (= (link cD) {cA cB cD}))
     (= (area) {cA cB cC cD}) (not (= (area) {cE cF}))
)
(:@\color{blue}\bfseries{global-goal}@ (and (= (pos p1) cA)(= (pos p3) cE)))
)
\end{lstlisting}}
\renewcommand{\ttdefault}{cmtt}

Sections of the problem file are also highlighted in bold. A domain file starts with a description of the {\ttfamily :objects} that the agent manages. As shown in the code, agents are represented as objects. Ag1 knows that there is a {\ttfamily truck t1} in the task, and it has knowledge of six different cities, although it only manages the four cities included in its working {\ttfamily area} (see Figure \ref{TransportTask}). The agent also knows that there are three packages in the MAP task, the {\ttfamily final-products p1} and {\ttfamily p2} and the {\ttfamily raw-material p3}.

The {\ttfamily :shared-data} section is a key aspect of our MAP language, as it defines the information shareable by the agents and directly affects their knowledge of the task. The predicates and functions defined in this structure are the patterns of the fluents that the agent regards as shareable with other agents. For instance, Ag1's {\ttfamily :shared-data} section includes the following pattern: {\ttfamily (empty ?c - city) - (either Ag2 Ag3)}. This pattern indicates that Ag1 will share the fluents that match the pattern {\ttfamily (empty ?c - city)} with both Ag2 and Ag3. Given that Ag1 knows the cities {\ttfamily cA}, {\ttfamily cB}, {\ttfamily cC}, {\ttfamily cD}, {\ttfamily cE} and {\ttfamily cF}, fluents as {\ttfamily (= (empty cA) true)} or {\ttfamily (= (empty cD) false)} match the pattern, and Ag1 shares this information with Ag2 and Ag3.

The {\ttfamily :init} section describes the initial state of Ag1, i.e., the initial situation of the world known to Ag1. It is defined with predicates like {\ttfamily (empty cB)}, functions like {\ttfamily (= (at t1) cA)}) and multi-functions like {\ttfamily (= (link cA) $\{$cB cC$\}$)}, that hold in the initial situation. The initial state includes both positive and negative information. For instance, the function {\ttfamily (not (= (at t1) cC))} indicates that Ag1 knows that {\ttfamily truck t1} is not initially placed at {\ttfamily city cC}. The information not included in the initial state is considered unknown to Ag1.

While the initial state contains predicates, functions and multi-functions, internally the system treats all of them as fluents. For instance, a predicate {\ttfamily (empty cB)} is internally converted into a fluent {\ttfamily (= (empty cB) true)}, while functions like {\ttfamily (= (at t1) cA)} are already in the form of fluents. Multi-functions are used to easily define multiple functions through a simplified notation. The conversion into fluents is straightforward: given a multi-function {\ttfamily (= (link cA) {cB cC})}, we generate the fluents {\ttfamily (= (link cA cB) true)} and {\ttfamily (= (link cA cC) true)}.

Finally, the {\ttfamily :global-goal} structure shows the global objective of the MAP task. In this case, the goal is to transport the {\ttfamily raw-material p3} to {\ttfamily city cF}, and to deliver a {\ttfamily final-product} to {\ttfamily city cA}. Notice that, in this example, there is not a {\ttfamily :private-goal} section.

\section{MAP algorithm overview}
\label{Planning_reasoning}

This section summarizes the main stages of the MAP algorithm followed by the agents to devise, exchange and select refinement plans to come up with a solution for the MAP task. Agents follow a procedure that integrates planning and coordination, allowing agents to solve both \emph{strongly-related} and \emph{loosely-coupled} problems without losing generality. Agents perform an individual Partial-Order Planning (POP) search to build refinements over the current base plan, while one of the agents leads the process of gathering the new refinement plans and selecting the next base plan.

\begin{algorithm}
\caption{Overview of the MAP algorithm}
\label{algorithm}
\begin{algorithmic}
\STATE Initial information exchange
\REPEAT
	\STATE Individual refinement process
	\STATE Coordination process
\UNTIL a solution plan is found or the search space is completely explored
\end{algorithmic}
\end{algorithm}

Algorithm \ref{algorithm} shows the main steps of the MAP algorithm. The stages of the algorithm are outlined as follows:

\begin{itemize}
	\item \textbf{Initial information exchange:} The algorithm starts with an initial communication stage by which agents exchange the shareable information on the planning domain in order to generate the data structures that will be used in the subsequent planning process. Agents take advantage of the exchanged information to build a distributed Relaxed Planning Graph, which provides them with their partial view on the MAP task.
	\item \textbf{Resolution process:} Once agents have exchanged the shareable information and the distributed Relaxed Planning Graph is computed, they start the iterative resolution process by which they explore the search space until they find a solution for the MAP task. As shown in Algorithm \ref{algorithm}, this process comprises two different interleaved stages, an individual planning process by which agents devise refinements over a centralized base plan and a coordination process to exchange the new refinement plans and to select the next base plan:
	\begin{itemize}
		\item \textbf{Individual refinement process:} Agents individually refine the current base plan of the MAP system. Each planning agent is provided with an internal POP system. The classical POP algorithm has been adapted to a MAP context in order to obtain valid refinement plans over an incomplete base plan (see section \ref{Refinement}).
		\item \textbf{Coordination process:} Agents communicate and exchange the new refinement plans over the current base plan. Later, they jointly evaluate these refinement plans and select the most promising one as the next base plan.
	\end{itemize}
\end{itemize}

The following sections detail the two main stages of the MAP algorithm. Section \ref{Initial} describes the initial information exchanging stage performed by the agents, while section \ref{Problem_solving} details the resolution process, including both the coordination process and the individual construction of the refinement plans.

\section{Initial information exchange}
\label{Initial}

Prior to the resolution process itself, agents perform a preliminary stage to share public planning information effectively. This initial stage builds a distributed Relaxed Planning Graph (dis-RPG), whose construction is inspired by the approach in \cite{Feng07}. Unlike the proposal in \cite{Feng07}, which stops the graph construction once all the problem goals appear in the graph, our procedure builds a complete dis-RPG by maintaining the \emph{incomplete information} of the agents, so they only exchange the information defined as shareable in the input files (see section \ref{Specification}). This section describes in detail the dis-RPG building process and subsection \ref{Example_RPG} provides a trace based on the MAP task presented in section \ref{M_example} that illustrates this process.

The dis-RPG provides the agents with valuable planning information that will be used throughout the refinement planning process:

\begin{itemize}
	\item Agents exchange the fluents defined as shareable in the {\ttfamily :shared-data} section of the MAP domain definition files (see subsection \ref{Shared_data}). Fluents are labeled with the list of agents that can achieve them, giving each agent a view of the possible interactions that can arise at planning time with other agents.
	\item An estimate of the best cost to achieve each fluent is computed. This information is used to design heuristics to guide the refinement planning process.
\end{itemize}

Following Algorithm \ref{RPG_algorithm}, the first step of the dis-RPG construction consists in computing the initial RPG for each agent $i$, $RPG^i$, taking only into account the fluents and actions initially known to the agent. Agents compute this initial planning graph by following the procedure presented in \cite{Hoffmann01}. The RPG consists of a set of alternating fluent and action levels. The first fluent level contains the fluents that are part of the initial state, and the first action level includes all the actions whose preconditions appear in the first fluent level. Fluents that are part of the effects of these actions (and have not been included in the first fluent level) are placed in the second fluent level, and actions whose preconditions are included in the two prior fluent levels of the graph (and are not in the first action level) are stored in the second action level. By following this procedure, the RPG is expanded until no new fluents are found. This way, the level of the graph in which an action or fluent appears gives an estimate of the cost of achieving such an action or fluent.

\begin{algorithm}
\caption{Dis-RPG construction for an agent $i$}
\label{RPG_algorithm}
\begin{algorithmic}
\STATE Build initial $RPG^i$
\REPEAT
	\STATE $\forall j \neq i$, $i$ sends $j$ shareable fluents $SF^{i\rightarrow j} \in RPG^i$ of the form $\langle v, d\rangle$ or $\langle v, \neg d\rangle$, where $v \in \Prop^i \cap \Prop^j$ and $d \in \D^{i}_v \cap \D^{j}_v$
	\STATE $\forall j \neq i$, $i$ receives from $j$ shareable fluents $SF^{j\rightarrow i} \in RPG^j$ of the form $\langle v, d\rangle$ or $\langle v, \neg d\rangle$, where $v \in \Prop^i \cap \Prop^j$ and $d \in \D^{i}_v \cap \D^{j}_v$
	\STATE $RF^i \gets \emptyset$
	\STATE $\forall j \neq i, RF^i \gets RF^i \cup SF^{j\rightarrow i}$
	\FORALL{received fluents $f \in RF^i$}
	\IF {$f \not\in RPG^i$}
		\STATE Insert f in $RPG^i$
        \STATE $level_{RPG^i}(f) \gets level(f)$
	\ENDIF
	\IF {$(f \in RPG^i)$ $\wedge$ $(level_{RPG^i}(f) > level(f))$}
		\STATE $level_{RPG^i}(f) \gets level(f)$
	\ENDIF
	\ENDFOR
	\STATE Expand $RPG^i$
\UNTIL $RF^i = \emptyset$
\end{algorithmic}
\end{algorithm}

Once all agents have computed their initial RPGs, the iterative composition of the dis-RPG begins. As depicted in Algorithm \ref{RPG_algorithm}, after computing the initial $RPG^i$, agent $i$ executes the first iteration of the algorithm and exchanges the fluents and actions of its $RPG^i$ with the rest of agents. Agents only exchange the fluents defined as shareable in the {\ttfamily :shared-data} structure of the input files (see subsection \ref{Shared_data}). Agent $i$ sends agent $j$ the set of fluents $SF^{i\rightarrow j}$ that are visible to agent $j$, i.e., the fluents in $RPG^i$ of the form $\langle v, d\rangle$ or $\langle v, \neg d\rangle$, where $v \in \Prop^i \cap \Prop^j$ and $d \in \D^{i}_v \cap \D^{j}_v$. Likewise, agent $i$ will receive from the rest of agents $j \neq i$ the shareable fluents of their $RPG^j$ that are visible to agent $i$.

Agent $i$ updates its $RPG^i$ accordingly with the new fluents received from the rest of agents. We will refer to these fluents as $RF^i$ (see Algorithm \ref{RPG_algorithm}). If a fluent $f \in RF^i$ is not in $RPG^i$ then it is stored according to $level(f)$. If $f$ is already in $RPG^i$, its level in the graph is updated if $level_{RPG^i}(f) > level(f)$. Hence, agents only store the best estimated level to reach each fluent, placing each fluent at the lowest possible level of the graph. After updating $RPG^i$, agent $i$ expands it by checking wether the new inserted fluents trigger new actions in $RPG^i$ or not. The fluents produced as effects of these new actions will be shared in the next information exchange iteration. The RPG expansion procedure also updates the existing actions by placing them at a lower action level if their preconditions have been updated.

Since agents only exchange those fluents defined as shareable, the dis-RPG process gives each agent a different view of the planning task, so no agent handles a complete representation of the dis-RPG. In contrast, each agent $i$ maintains its own internal $RPG^i$, whose information depends on the fluents other agents have shared with it, which makes each agent have its own, partial view of the planning task. Thus, agents design plans under \emph{incomplete information}, as they are partly aware of the information on the planning task.
\begin{table}
\begin{scriptsize}
\begin{tabular}{|l|l|}
\hhline{--}
F0 & \\
\hhline{--}
$[{\color{red}2}] $(empty cB) T & $[{\color{red}2}]$(link cE cB) T \\
$[{\color{red}2}]$(empty cD) T & $[{\color{red}2}]$(link cE cD) T \\
$[{\color{red}2}]$(empty cE) T & $[{\color{red}2}]$(link cF cD) T \\
$[{\color{red}2}]$(empty cF) T & $[{\color{red}2}]$(area cB) T \\
$[{\color{red}2}]$(link cB cE) T & $[{\color{red}2}]$(area cD) T \\
$[{\color{red}2}]$(link cD cE) T & $[{\color{red}2}]$(area cE) T \\
$[{\color{red}2}]$(link cD cF) T & $[{\color{red}2}]$(area cF) T \\
\hhline{--}
\end{tabular}
\end{scriptsize}
\caption{Initial RPG built by agent Ag2}
\label{RPG2I}
\end{table}

\begin{table}
\begin{scriptsize}
\begin{tabular}{|l|l|l|l|l|}
\hhline{-----}
F0 & & A0 & F1 & A1\\
\hhline{-----}
$[{\color{blue}1}]${\color{red}(at t1) cA} & $[{\color{blue}1}]${\color{red}(pos p3) cA} & load t1 p3 cA &$[{\color{blue}1}]${\color{red}(pos p3) t1} & unload t1 p3 cB\\
$[{\color{blue}1}]${\color{red}(empty cA) F} & $[{\color{blue}1}]$(link cC cA) T & drive t1 cA cB & $[{\color{blue}1}]${\color{red}(empty cA) T} & unload t1 p3 cC\\
$[{\color{blue}1}]${\color{red}(empty cB) T} & $[{\color{blue}1}]$(link cC cB) T & drive t1 cA cC & $[{\color{blue}1}]${\color{red}(at t1) cB} & unload t1 p3 cA\\
$[{\color{blue}1}]${\color{red}(empty cC) T} & $[{\color{blue}1}]$(link cC cD) T & & $[{\color{blue}1}]${\color{red}(at t1) cC} & drive t1 cB cA\\
$[{\color{blue}1}]${\color{red}(empty cD) T} & $[{\color{blue}1}]$(link cD cC) T & & & drive t1 cB cC\\
$[{\color{blue}1}]$(link cA cB) T & $[{\color{blue}1}]$(area cA) T & & & drive t1 cC cA\\
$[{\color{blue}1}]$(link cA cC) T & $[{\color{blue}1}]$(area cB) T & & & drive t1 cC cB\\
$[{\color{blue}1}]$(link cB cA) T & $[{\color{blue}1}]$(area cC) T & & & drive t1 cC cD\\
$[{\color{blue}1}]$(link cB cC) T & $[{\color{blue}1}]$(area cD) T & &\\
\hhline{-----}
\end{tabular}

\medskip

\begin{tabular}{|l|l|l|l|}
\hhline{----}
F2 & A2 & F3 & A3\\
\hhline{----}
$[{\color{blue}1}]${\color{red}(empty cB) F} & load t1 p3 cB & $[{\color{blue}1}]${\color{red}(empty cD) F} & load t1 p3 cD \\
$[{\color{blue}1}]${\color{red}(empty cC) F} & load t1 p3 cC & $[{\color{blue}1}]${\color{red}(pos p3) cD} & \\
$[{\color{blue}1}]${\color{red}(at t1) cD} &	unload t1 p3 cD & & \\
$[{\color{blue}1}]${\color{red}(pos p3) cB} & drive t1 cD cC & & \\
$[{\color{blue}1}]${\color{red}(pos p3) cC} & & & \\
\hhline{----}
\end{tabular}

\end{scriptsize}
\caption{Initial RPG built by agent Ag1}
\label{RPG1I}
\end{table}

The dis-RPG process finishes when agents do not receive more fluents from the others. Following, agents start the resolution process to jointly devise a solution plan.

\subsection{dis-RPG example}
\label{Example_RPG}

In order to illustrate the dis-RPG stage of the MAP algorithm, this section provides an example trace based on the transportation and storage MAP task introduced in section \ref{M_example}. The planning agents receive the input files presented in subsection \ref{Example_modeling} and start the MAP algorithm by building the dis-RPG.

In the first stage of Algorithm \ref{RPG_algorithm}, each agent individually generates an initial RPG, according to its problem file. To illustrate this stage of the process, we focus on the initial RPGs built by the transport agents Ag1 and Ag2.

Table \ref{RPG2I} shows the initial RPG calculated by agent Ag2. The numbers in brackets indicate the agents that can generate the fluent, while the values \emph{T} and \emph{F} stand for \emph{true} and \emph{false}, respectively. Ag2 does not know the position of the packages and the truck because they are initially located out of its working area (see Figure \ref{TransportTask} in section \ref{M_example}). Therefore, its initial RPG only includes F0, the first level of fluents, which stores the fluents on the initial state of Ag2. The initial RPG of Ag2 does not contain any action level because there are no applicable actions, that is, their preconditions do not hold in F0.

Agent Ag1 does know the position of the package {\ttfamily p3} and the truck  {\ttfamily t1}, and consequently, it can compute a much larger initial RPG (see Table \ref{RPG1I}). Notice that the level A0 includes the actions whose preconditions are satisfied in F0, while F1 stores the fluents that are part of the effects of the actions in A0 and are not in F0. For instance, the action {\ttfamily drive t1 cA cB}, at level A0, has the following preconditions: {\ttfamily (= (area) cA)}, {\ttfamily (= (area) cB)}, {\ttfamily (= (link cA cB) true)} and {\ttfamily (= (at t1) cA)}. As Table \ref{RPG1I} shows, these fluents are at F0, which triggers the action {\ttfamily drive t1 cA cB} at A0.

\begin{table}
\begin{scriptsize}
\begin{tabular}{|l|l|l|l|l|}
\hhline{-----}
F0 & & A0 & F1 & A1 \\
\hhline{-----}
$[{\color{blue}1},{\color{red}2}] $(empty cB) T & $[{\color{red}2}]$(link cE cB) T & load t1 p3 cA & $[{\color{blue}1}]$(empty cA) T & unload t1 p3 cB\\
$[{\color{blue}1},{\color{red}2}]$(empty cD) T & $[{\color{red}2}]$(link cE cD) T & drive t1 cA cB & $[{\color{blue}1},{\color{red}2}]$(at t1) cB & unload t1 p3 cC\\
$[{\color{red}2}]$(empty cE) T & $[{\color{red}2}]$(link cF cD) T & drive t1 cA cC & $[{\color{blue}1}]$(at t1) cC & unload t1 p3 cA\\
$[{\color{red}2}]$(empty cF) T & $[{\color{red}2}]$(area cB) T & & $[{\color{blue}1},{\color{red}2}]$(pos p3) t1 & drive t1 cB cA\\
$[{\color{red}2}]$(link cB cE) T & $[{\color{red}2}]$(area cD) T & & & drive t1 cB cC\\
$[{\color{red}2}]$(link cD cE) T & $[{\color{red}2}]$(area cE) T & & & drive t1 cC cA\\
$[{\color{red}2}]$(link cD cF) T & $[{\color{red}2}]$(area cF) T & & & drive t1 cC cB\\
$[{\color{blue}1}]$(empty cA) F & $[{\color{blue}1}]$(at t1) cA & & & drive t1 cC cD\\
$[{\color{blue}1}]$(pos p3) cA & $[{\color{red}2},{\color{darkgreen}3}]$(empty cF) T & & \\
\hhline{-----}
\end{tabular}

\medskip

\begin{tabular}{|l|l|l|l|}
\hhline{----}
F2 & A2 & F3 & A3\\
\hhline{----}
$[{\color{blue}1},{\color{red}2}]$(empty cB) F & load t1 p3 cB & $[{\color{blue}1},{\color{red}2}]$(empty cD) F & load t1 p3 cD \\
$[{\color{blue}1}]$(empty cC) F &	load t1 p3 cC & $[{\color{red}2}]$(at t1) cF & \\
$[{\color{blue}1},{\color{red}2}]$(at t1) cD & unload t1 p3 cD & $[{\color{blue}1},{\color{red}2}]$(pos p3) cD & \\
$[{\color{red}2}]$(at t1) cE & drive t1 cD cC & $[{\color{red}2}]$(pos p3) cE & \\
$[{\color{blue}1},{\color{red}2}]$(pos p3) cB & & $[{\color{red}2}]$(empty cE) F & \\
$[{\color{blue}1}]$(pos p3) cC & & $[{\color{red}2},{\color{darkgreen}3}]$(pos p2) cF & \\
$[{\color{red}2},{\color{darkgreen}3}]$(pos p1) cF & & & \\
$[{\color{blue}1},{\color{red}2}]$(empty cF) F & & & \\
\hhline{----}
\end{tabular}

\medskip

\begin{tabular}{|l|l|l|l|l|l|}
\hhline{----}
F4 & A4 & F5 & A5\\
\hhline{----}
$[{\color{red}2}]$\textbf{(pos p3) cF} & unload t1 p1 cB & $[{\color{blue}1}]$\textbf{(pos p1) cA} & load t1 p1 cB \\
$[{\color{blue}1},{\color{red}2}]$(pos p1) t1 & unload t1 p1 cC & $[{\color{blue}1},{\color{red}2}]$(pos p1) cB & load t1 p1 cD \\
$[{\color{blue}1},{\color{red}2}]$(pos p2) t1 & unload t1 p1 cD & $[{\color{blue}1}]$(pos p1) cC & load t1 p2 cB\\
 & unload t1 p1 cA & $[{\color{blue}1},{\color{red}2}]$(pos p1) cD & load t1 p2 cD \\
 & unload t1 p2 cB & $[{\color{red}2}]$(pos p1) cE & load t1 p1 cC \\
 & unload t1 p2 cC & $[{\color{blue}1}]$(pos p2) cA & load t1 p1 cA \\
 & unload t1 p2 cD & $[{\color{blue}1},{\color{red}2}]$(pos p2) cB & load t1 p2 cC \\
 & unload t1 p2 cA & $[{\color{blue}1}]$(pos p2) cC & load t1 p2 cA \\
 & & $[{\color{blue}1},{\color{red}2}]$(pos p2) cD & \\
 & & $[{\color{red}2}]$(pos p2) cE & \\
\hhline{----}
\end{tabular}
\end{scriptsize}
\caption{Final dis-RPG as viewed by agent Ag2}
\label{dis-RPG}
\end{table}

Once agents have built their initial RPGs, they start the iterative dis-RPG building process by exchanging the shareable fluents in their RPGs.

In subsection \ref{pfiles}, we show the {\ttfamily :shared-data} section of Ag1, which shares with Ag2 fluents that match the following patterns: {\ttfamily (empty ?c - city)}, {\ttfamily ((at ?t - truck) - city)} and {\ttfamily ((at ?t - truck) - city)}. The fluents shared by Ag1 and Ag2 are marked in red in Table \ref{RPG1I}. Ag2 also sends its shareable fluents to the rest of agents and stores the fluents received from other agents.

Agents expand their RPGs by checking if the fluents they have received trigger new actions in the graph. The process carries on until no new fluents appear in the dis-RPG. As each agent has a different {\ttfamily :shared-data} section, the information will vary from one RPG to another, giving each agent a different and incomplete view of the dis-RPG and the MAP task itself.

Table \ref{dis-RPG} shows the final dis-RPG of the transportation scenario as seen by agent Ag2. As it can be observed, the dis-RPG provides both an estimate of the cost of achieving each fluent (this cost corresponds to the level at which the fluent appears), and the set of agents that achieve that fluent in the RPG.
\vspace{-0.1cm}
\section{Resolution process}
\label{Problem_solving}

After the information exchange, agents initiate the resolution process (see Algorithm \ref{Problem_solving_algorithm}), which comprises two interleaved stages: the individual refinement stage and the coordination stage. The first stage involves agents building individual refinements over a centralized base plan by using a POP. In the second stage, agents follow a coordination process to gradually build a joint solution plan for the MAP task, exchanging and evaluating the refinements generated individually and selecting the most promising one in order to reach a solution.

\begin{algorithm}
\caption{Resolution process for an agent $i$}
\label{Problem_solving_algorithm}
\begin{algorithmic}
\STATE $\Pi \gets \Pi_0$
\STATE $R = \emptyset$
\REPEAT
	\STATE Select open goal $g \in \openGoals(\Pi)$
	\STATE $Refinements^i(\Pi_{g}) \gets$ Refine base plan $\Pi_{g}$ individually
    \STATE $\forall j \not= i$, send $Refinements^i(\Pi_{g})$ to agent $j$
    \STATE $\forall j \not= i$, receive $Refinements^j(\Pi_{g})$
    \STATE $Refinements(\Pi_{g}) \gets Refinements^i(\Pi_{g})$
    \STATE $\forall j \not= i$, $Refinements(\Pi_{g}) \gets Refinements(\Pi_{g}) \cup Refinements^j(\Pi_{g})$
	\STATE Evaluate $Refinements(\Pi_{g})$
	\STATE $R \gets R \cup Refinements(\Pi_{g})$
	\STATE Vote for the best plan $\Pi^i \in R$
	\STATE $\Pi \gets \Pi^i$
	\IF {$\openGoals(\Pi) = \emptyset$}
		\RETURN $\Pi$
	\ENDIF
\UNTIL $R = \emptyset$
\end{algorithmic}
\end{algorithm}
\vspace{-0.3cm}
\subsection{Individual refinement stage}
\label{Refinement}

A planning agent $i$ executes its individual POP process in order to refine the current base plan $\Pi$. As shown in Algorithm \ref{Problem_solving_algorithm}, agent $i$ refines $\Pi$ by solving a particular open goal $g \in \openGoals(\Pi)$, thus obtaining a set of valid refinement plans over $\Pi_{g}$, $Refinements^i(\Pi_{g})$.

Our definition of refinement plan (see subsection \ref{MAPOP}) states that a refinement plan $\Pi^i$ of an agent $i$ over a base plan $\Pi$ solves one of its open goals $g \in \openGoals(\Pi)$, as well as all the private open goals $g^i$ of the form $\langle v, d\rangle$ or $\langle v, \neg d\rangle$ that arise from this resolution, where $v \in \Prop^i \wedge d \in \D^{i}_v \wedge ((\forall j \not= i, v \not\in \Prop^j) \vee (\forall j \not= i, d \not\in \D^{j}_v)) \wedge (g^i \not\in \openGoals(\Pi))$.

We have designed a customized version of the classical POP algorithm compliant with the requirements introduced by the MAP approach. Our POP system is able to start the search process from any given base plan, rather than starting with an empty plan as in a traditional POP process. In addition, the POP is aimed at building refinement plans, rather than complete solution plans.

\subsection{Coordination process}

The coordination process is based on a democratic leadership in which a leadership baton is scheduled among the agents (initially, the baton is randomly assigned to one of the participating agents). The resolution process interleaves the coordination process with the individual refinement stage. A coordination iteration is led by the agent which has the baton (baton agent). Once the coordination iteration is completed, the baton is handed over to the following agent.

Algorithm \ref{Problem_solving_algorithm} depicts the main steps of the coordination process. After the individual refinement stage, agents exchange the refinement plans they have elicited over the current base plan $\Pi$. Following, agents receive the refinement plans of the other agents and evaluate them according to their view of the planning task. Agents apply a voting process to adopt the most promising plan as the next base plan $\Pi$, and check if the selected plan is a solution. Otherwise, agents choose a new open goal of the plan $g \in \openGoals(\Pi)$ and each agent $i$ starts a new individual refinement stage to compute the refinements over $\Pi$, $Refinements^i(\Pi_g)$.

In the first step of the coordination process, the individual refinement plans are exchanged between agents for their evaluation. An agent $i$ sends the refinement plans it has devised over the current base plan $\Pi$ by solving $g \in \openGoals(\Pi)$, $Refinements^i(\Pi_g)$, to the rest of agents in the task. In turn, agent $i$ receives the refinements devised by each agent $j$ in the task, $Refinements^j$ $(\Pi_g)$, where $j \neq i$. Note that agents have a local, partial view of the plans, so given a refinement plan $\Pi$, an agent $i$ will only view the open goals $og \in \openGoals(\Pi)$ of the form $\langle v, d\rangle$ or $\langle v, \neg d\rangle$ such that $v \in \Prop^i$ and $d \in \D^{i}_v$. The view agent $i$ has on each refinement plan $\Pi$, $view^i(\Pi)$, ensures agents' privacy and directly affects the evaluation of the refinements.

The evaluation of the refinement plans is carried out through a utility function $\F$, by which agents  estimate the quality of the plans. Since an agent $i$ evaluates a plan accordingly to its view, $\F(view^i(\Pi))$, the results of the evaluation may be different from the other agents'. Therefore, agents will have different perspectives on the quality of the refinement plans.

Once the refinement plans are evaluated, agents vote for the most promising candidate in $R$, which stores all the refinement plans that have not yet been selected as a base plan (see Algorithm \ref{Problem_solving_algorithm}). Each agent $i$ votes for the best refinement plan in $R$ according to the utility function $\F$. In case of a draw, the baton agent will choose the next base plan among the most voted alternatives. If $R = \emptyset$, the refinement planning process ends with no solution found.

Once a refinement plan is selected, agents adopt it as the new base plan $\Pi$. If $\openGoals(\Pi) = \emptyset$, a solution plan is returned. As some open goals might not be visible to some agents, all agents must confirm that $\Pi$ is a solution plan according to their view of $\Pi$. Finally, the baton agent selects the next open goal $g \in \openGoals(\Pi)$ to be solved, and a new iteration of the refinement and coordination process starts.

The resolution process carried out by the agents can be regarded as a joint exploration of the refinement space. Nodes in the search tree represent refinement plans and each iteration of the process expands a different node.

\subsection{Resolution example}
\label{Example_resolution}

This subsection illustrates the resolution process by showing a partial trace that follows the trace example described in subsection \ref{Example_RPG}. After completing the initial information exchange and building the dis-RPG, agents proceed with the resolution process in order to solve the MAP task.

The plan construction starts with an initial empty plan, $\Pi_0$, which contains only the two fictitious steps that represent the initial state and the goals of the MAP task. The first open goal selected by Ag1 (which takes the role of baton agent in this first iteration) for its resolution is {\ttfamily (= (pos p1) cA)}, as it is the most costly one according to the dis-RPG. The goals of the task are highlighted in bold in Table \ref{dis-RPG}. This dis-RPG shows that {\ttfamily (= (pos p1) cA)} has a cost of 5, {\ttfamily (= (pos p3) cF)} has a cost of 4, and the only agent that can achieve {\ttfamily (= (pos p1) cA)} is Ag1.  Hence, Ag1 proposes a set of refinements over $\Pi_0$, $\Pi_{00}, \dots, \Pi_{09}$, while Ag2 and Ag3 refrain from making proposals. The proposed refinements are evaluated through the utility function $\F$, and the best-valued one, $\Pi_{00}$, is selected as the new base plan.

\begin{figure}
\centering
\includegraphics[width=10cm]{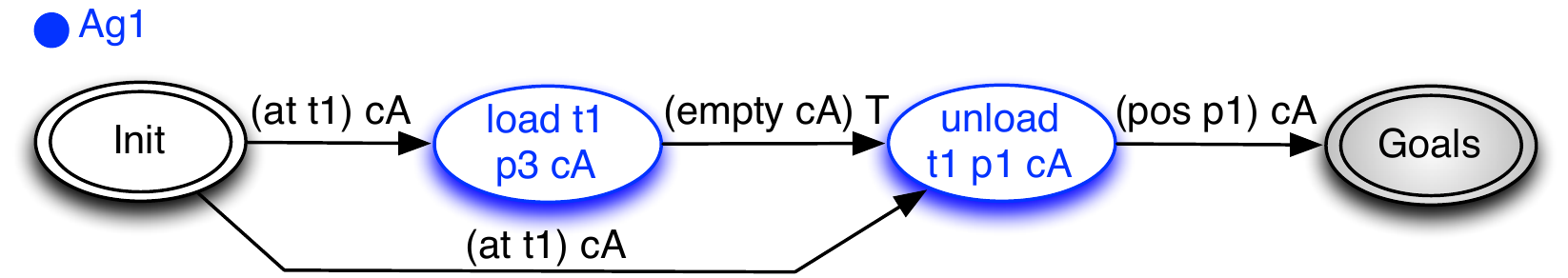}
\caption{Refinement plan $\Pi_{00}$}
\label{Refinement00}
\end{figure}

\begin{figure}
\centering
\includegraphics[width=12cm]{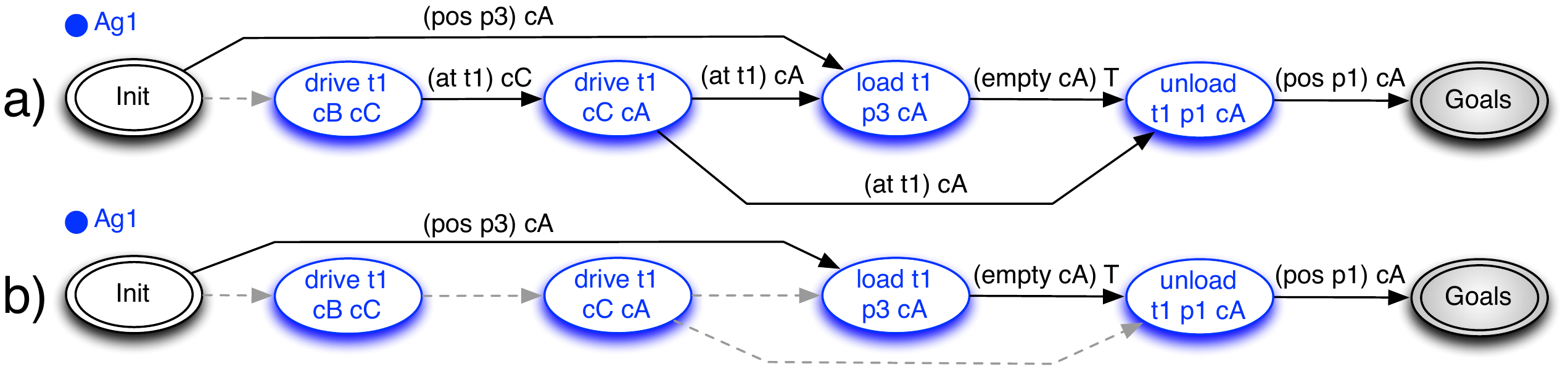}
\caption{Refinement plan $\Pi_{06}$ as observed by: a) Ag1 b) Ag3}
\label{Refinement06}
\end{figure}

Figure \ref{Refinement00} depicts the refinement plan $\Pi_{00}$. Since all the causal links in $\Pi_{00}$ involve shareable fluents, all the agents have a complete view of this refinement plan. However, agents Ag1 and Ag3 have different views of the refinement $\Pi_{06}$ (see Figure \ref{Refinement06}). In order to guarantee privacy, several causal links (black arrows) of $\Pi_{06}$ have been occluded to Ag3, which only sees ordering constraints instead (grey arrows). According to the problem definition files, the fluents involved in these causal links are private to the transport agent Ag1 because they are not shareable data, and therefore, Ag1 does not communicate them to Ag3.

\begin{figure}
\centering
\includegraphics[width=12.5cm]{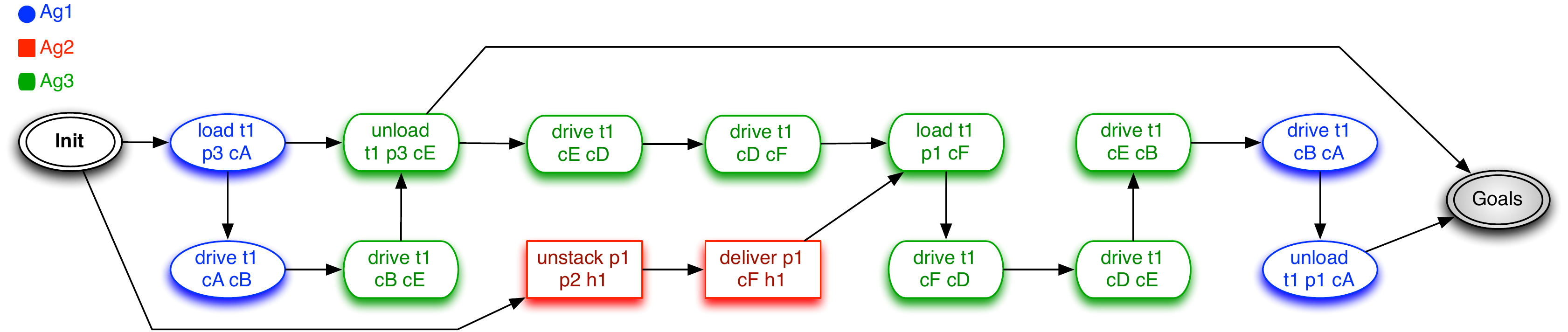}
\caption{Solution plan for the MAP task}
\label{Solution}
\end{figure}

Once the refinement plan $\Pi_{00}$ is chosen as the new base plan, Ag1 passes on the baton to Ag2 and a new iteration of the resolution process starts. The MAP process will carry on until a solution plan is found. Since some open goals are not visible to some agents, all participating agents must confirm that the plan has no pending open goals. Figure \ref{Solution} depicts the solution plan for the MAP task at hand, showing in different shapes the actions to be executed by each agent. As it can be observed, the solution of the MAP task is a joint plan to which all the participant agents have contributed.

\section{Experimental results}
\label{Results}

Several tests have been performed to evaluate the performance of our MAP system. The tests compare the MAP model with a single-agent approach to analyze its advantages and shortcomings against a centralized planning model. We have used two different planning domains for our experiments. Next subsections present the MAP domains and analyze the results of the different tests.

\subsection{Multi-agent planning domains}
\label{Domains}

The two planning domains used to test the MAP system have been taken from real-life problems or adapted from well-known case studies. The two domains were designed such that we could test the performance and the quality of the solutions obtained with a wide range of problems. We tested our MAP system with different levels of complexity: from \emph{loosely-coupled} problems, in which interactions among agents are rather low, to \emph{strongly-related} problems, that require a strong coordination effort to be solved. Additionally, we created both a multi-agent and a single-agent version for each problem.

In section \ref{M_example}, we described a transportation and storage domain, in which agents take the roles of transport agencies and storage facilities, which work together to transport raw materials and final products to different cities. This domain gives rise to \emph{strongly-related} problems as interactions between agents are required in order to accomplish the different objectives. Agents in the \emph{transportation} domain have different abilities, so they should cooperate with each other in order to achieve the different goals.

We defined an additional planning domain, the \emph{picture} domain. This domain gives rise to simpler, \emph{loosely-coupled} problems as agents can work independently in order to solve the objectives, and hence cooperation and interactions among agents are not mandatory to find a solution. Planning agents in the picture domain (workers) are not specialized, they all share the same abilities and so they all can perform the same actions. In addition, agents in this domain do not keep private information for themselves but all the problem information is shared among the agents. Next subsection describes the \emph{picture} domain.

\subsubsection{Picture domain}
\label{picture_dom}

This domain, adapted from the case study in \cite{Parsons98}, presents a situation in which several workers have to cooperate to hang a set of pictures on walls. To do so, they have to acquire different tools that are scattered over several locations. Agents move through the locations to get the tools and hang the pictures. The domain defines a set of bidirectional links that connect the locations.

\begin{figure}
\centering
\includegraphics[width=7.5cm]{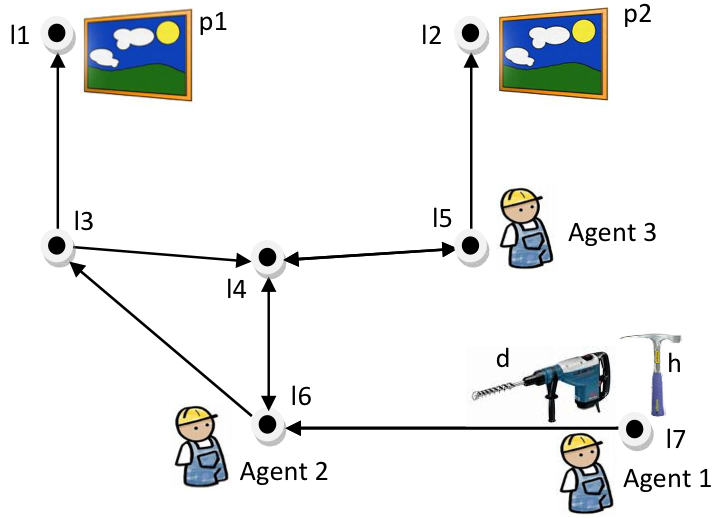}
\caption{Picture domain example}
\label{PictureScenario}
\end{figure}

Figure \ref{PictureScenario} depicts an example of this planning domain. In contrast to the transportation domain, agents in the picture domain share the same capabilities: agents can {\ttfamily pickUp} and {\ttfamily putDown} a {\ttfamily tool} in the {\ttfamily location} where the {\ttfamily agent} and the {\ttfamily tool} are placed; an agent can also {\ttfamily pass} the {\ttfamily tool} it is carrying on to another {\ttfamily agent} at the same {\ttfamily location}; agents can {\ttfamily walk} from one {\ttfamily location} to another through the link that connects both locations; finally, an agent can {\ttfamily hang} a {\ttfamily picture} on a certain {\ttfamily location} with the {\ttfamily tool} it is carrying.

This domain gives rise to \emph{loosely-coupled} problems because an individual agent is likely to solve the problem goals by itself in most cases. Moreover, agents share the same abilities and have access to all the locations, so they are able to work independently and cooperation is not a requirement to complete the task. Cooperation is however useful to improve the quality of the solutions and to solve conflicts on the use of the tools, as they are limited resources shared by all the participating agents.

\subsection{Tests and results}

The following subsections show the experimental results. We carried out two different tests\footnote{All the tests were performed on a single machine with a 2.83 GHz Intel Core 2 Quad CPU and 8 GB RAM.}. The first one compares the quality of the solution plans obtained by a single agent and by a set of planning agents working together on the problem. To do so, we defined a set of MAP problems and the single-agent equivalent version. Finally, we measured the robustness and scalability of the MAP system by executing a planning problem several times, increasing each time the number of planning agents in the system.
%\vspace{-2cm}
\subsubsection{Multi-Agent vs. Single-Agent Planning}

This first set of tests compares the quality of the solution plans of our MAP approach versus the centralized single-agent framework. The testbed includes twenty planning problems (ten problems per domain) of increasing difficulty.

As stated in subsection \ref{Domains}, the \emph{transportation} problems present a high coupling level  because agents are required to interact between each other to solve most of the problem goals. In contrast, agents in the \emph{picture} problems can solve the goals independently in most cases so the coupling level in these problems is rather low. Another key difference between both domains is that planning agents in the \emph{picture} domain (workers) are also the entities that execute the plans, whereas agents in the \emph{transportation} domain are merely planning entities. This way, given two parallel actions in a plan of the picture domain, each one is associated to a different agent (worker) whereas two parallel actions in a plan of the transportation domain can be associated to two different trucks of the same transport agent, which is the planning entity. In other words, concurrency is associated to the agents in the \emph{picture} domain and to the resources managed by the agents (trucks, hoists, etc.) in the \emph{transportation} domain.

Table \ref{Res_MAPSAP} shows the obtained results. $\#$Ag indicates the number of agents that perform the planning problem in the MAP tests. $\#$Actions and $\#$TS refer to the number of actions and \textit{time steps} of the solution plan, respectively (notice that we do not count the plans' fictitious actions). Finally, Parallelism indicates the maximum number of parallel branches in the MAP solution plans.

\begin{figure}
\centering
\includegraphics[width=12.2cm]{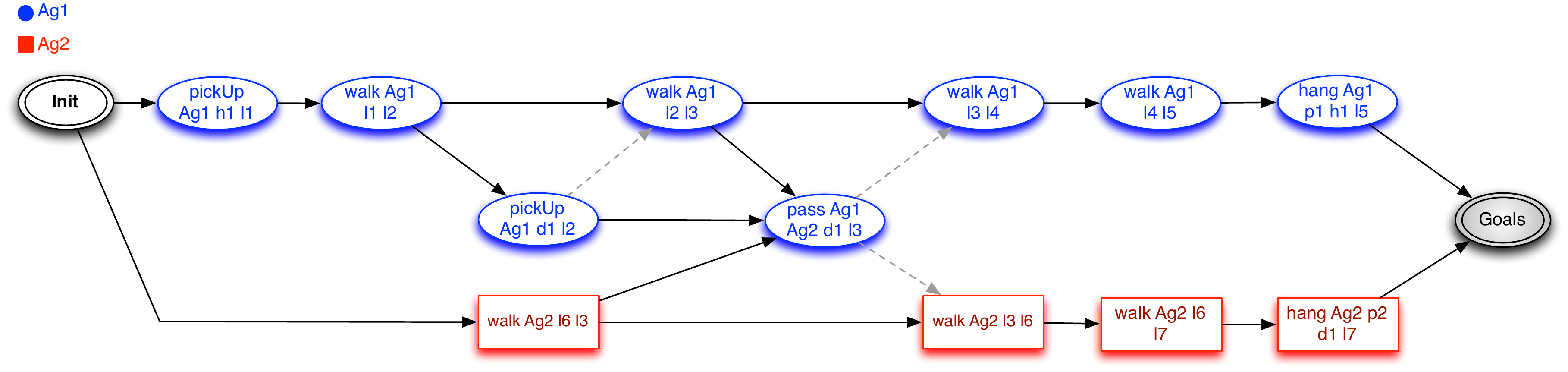}
\caption{Solution plan for the Picture2 MAP problem}
\label{Plan}
\end{figure}

Time steps are the number of time units necessary to execute the plan, i.e., the duration of the plan. For instance, Figure \ref{Plan} depicts the solution plan for the Picture2 MAP problem. Although the plan is composed of twelve planning actions (without taking into account the two fictitious actions), it can be executed in only eight time steps, since most of its actions can be executed in two parallel branches. Then, the duration of the plan in Figure \ref{Plan} is 8 time units.
%\vspace{-0.3cm}
\paragraph{Discussion on the results.}

In the \emph{transportation} domain, the MAP approach obtains the same results than the single-agent approach w.r.t. the number of actions and time steps. The single-agent approach performs rather well in this particular domain, obtaining good-quality solutions, if not optimal, for almost all the tested problems. Notice that the single-agent approach features a single planning entity that has a full visibility on the planning problem. Despite the fact that the participating agents on the MAP tests have an incomplete view of the problem, the results show that MAP agents cooperate effectively, obtaining plans of the same quality as the single-agent approach, both in terms of number of actions and plan duration (time steps).

\begin{center}
\begin{table}
\begin{tabular}{ | c || r | r | r | r || r | r |}
	\hhline{-------}
    \multicolumn{1}{|c||}{\multirow{2}{*}{Problem}} &
    \multicolumn{4}{c||}{Multi-Agent Planning} &
    \multicolumn{2}{c|}{Single-Agent Planning} \\ \cline{2-7}
    \multicolumn{1}{|c||}{} &
    \multicolumn{1}{c|}{$\#$Ag} &
    \multicolumn{1}{c|}{$\#$Actions} &
    \multicolumn{1}{c||}{$\#$TS} &
    \multicolumn{1}{c||}{Parallelism} &
    \multicolumn{1}{c|}{$\#$Actions} &
    \multicolumn{1}{c|}{$\#$TS} \\ \hhline{=======}
    \multicolumn{1}{|c||}{Transportation1} & 2 & 14 & 11 & 2 & 14 & 11 \\ \hhline{-------}
    \multicolumn{1}{|c||}{Transportation2} & 2 & 11 & 9 & 2 & 11 & 9 \\ \hhline{-------}
    \multicolumn{1}{|c||}{Transportation3} & 3 & 9 & 5 & 2 & 9 & 5 \\ \hhline{-------}
    \multicolumn{1}{|c||}{Transportation4} & 3 & 11 & 6 & 2 & 11 & 6 \\ \hhline{-------}
    \multicolumn{1}{|c||}{Transportation5} & 4 & 13 & 6 & 3 & 13 & 6 \\ \hhline{-------}
    \multicolumn{1}{|c||}{Transportation6} & 4 & 11 & 5 & 3 & 11 & 5 \\ \hhline{-------}
    \multicolumn{1}{|c||}{Transportation7} & 5 & 10 & 8 & 2 & 10 & 8 \\ \hhline{-------}
    \multicolumn{1}{|c||}{Transportation8} & 5 & 15 & 9 & 3 & 15 & 9 \\ \hhline{-------}
    \multicolumn{1}{|c||}{Transportation9} & 6 & 11 & 5 & 3 & 11 & 5 \\ \hhline{-------}
    \multicolumn{1}{|c||}{Transportation10} & 6 & 17 & 10 & 3 & 17 & 10 \\ \hhline{=======}

    \multicolumn{1}{|c||}{Picture1} & 2 & 11 & 6 & 2 & 14 & 14 \\ \hhline{-------}
    \multicolumn{1}{|c||}{Picture2} & 2 & 12 & 8 & 2 & 11 & 11 \\ \hhline{-------}
    \multicolumn{1}{|c||}{Picture3} & 3 & 6 & 2 & 3 & 8 & 8 \\ \hhline{-------}
    \multicolumn{1}{|c||}{Picture4} & 3 & 11 & 7 & 2 & 11 & 11 \\ \hhline{-------}
    \multicolumn{1}{|c||}{Picture5} & 4 & 8 & 2 & 4 & 11 & 11 \\ \hhline{-------}
    \multicolumn{1}{|c||}{Picture6} & 4 & 10 & 6 & 2 & 10 & 10 \\ \hhline{-------}
    \multicolumn{1}{|c||}{Picture7} & 5 & 8 & 5 & 2 & 8 & 8 \\ \hhline{-------}
    \multicolumn{1}{|c||}{Picture8} & 5 & 10 & 2 & 5 & 14 & 14 \\ \hhline{-------}
    \multicolumn{1}{|c||}{Picture9} & 6 & 9 & 5 & 2 & 9 & 9 \\ \hhline{-------}
    \multicolumn{1}{|c||}{Picture10} & 6 & 12 & 2 & 6 & 17 & 17 \\ \hhline{-------}
\end{tabular}
\caption{Single-Agent vs. Multi-Agent Planning comparison}
\label{Res_MAPSAP}
\end{table}
\end{center}

\vspace{-0.85 cm}
In the \emph{transportation} domain, planning agents have a set of resources at their disposal (truck and hoists) to execute the actions of the plan. Since partial-order planners allow for parallelism, both the MAP and single-agent plans contain parallel actions. Actions in this domain are executed by the trucks and hoists instead of the planning agents themselves. Hence, the number of parallel branches and the duration of the solution plans of this domain is only conditioned by the number of available resources (trucks and hoists). For this reason, both approaches give rise to plans with the same number of actions and time steps. On the basis of these results, we can affirm that the quality of the MAP plans is not diminished by the limited view and \emph{incomplete information} of the agents and the existence of private information among agents.

The results of the \emph{picture} domain present more differences between both approaches. The single-agent approach obtains sequential plans because the single planning agent is also the only execution entity. MAP, however, takes advantage of having several planning/execution agents cooperating. MAP enforces cooperation as agents can work together to reach an objective. For instance, Figure \ref{Plan} shows that an agent can pick up a tool and pass it on to another agent. This cooperation improves the solution because it prevents the agent from going for the tool and retrace its steps, thus reducing the number of actions of the plan. Agents also cooperate by proposing different parts of the plan that can be executed concurrently, which reduces the duration of the plans with respect to the centralized approach. Table \ref{Res_MAPSAP} shows that all the MAP solution plans for the \emph{picture} domain include at least two parallel branches of actions, meaning that at least two agents work concurrently, which improves the quality of the solutions as shown in Table \ref{Res_MAPSAP}.

In conclusion, while being a more costly approach (see next subsection for scalability tests), MAP obtains equal or better solution plans in terms of both number of actions and duration of the plans than the single-agent model. We have shown that MAP promotes cooperation among agents thus improving the quality of the solution. In addition, MAP agents manage their \emph{incomplete information} on the MAP task efficiently as the quality of the solution plans is not affected, being at least on par with the single-agent approach. Moreover, results show that our approach obtains good-quality solution plans for problems with different coupling and complexity levels, from \emph{loosely-coupled} to \emph{strongly-related} problems.

\begin{figure}
\centering
\includegraphics[width=12cm]{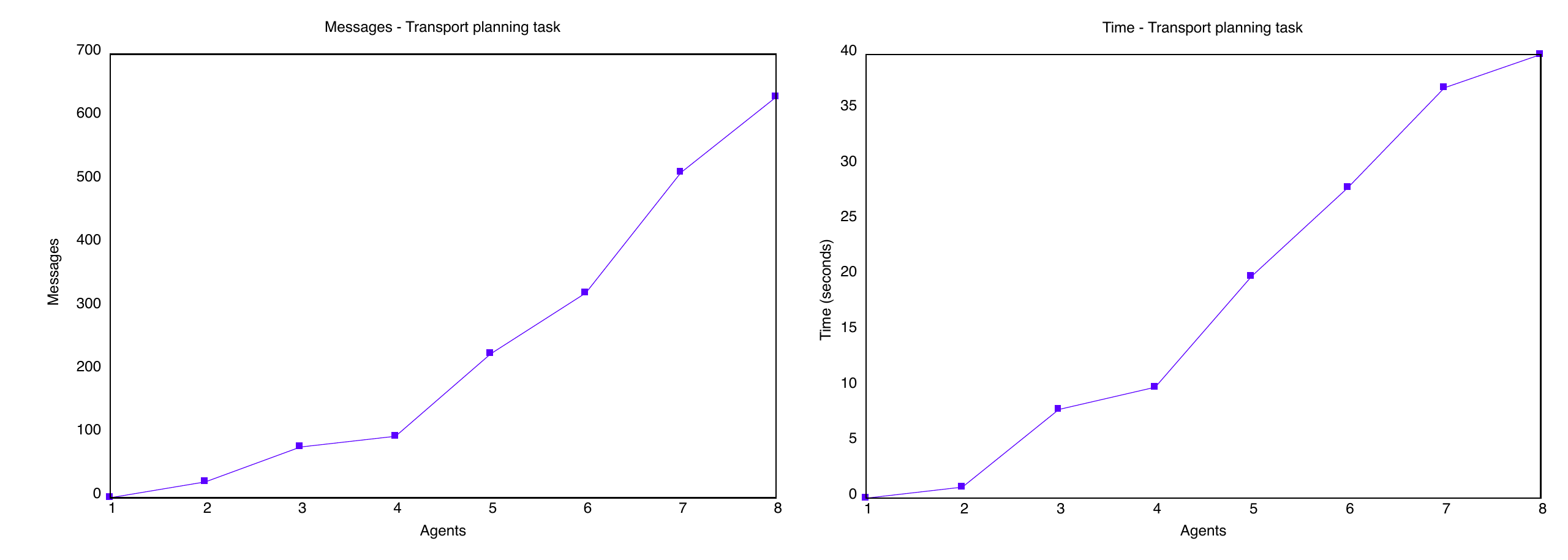}
\caption{Scalability results for the transportation domain}
\label{GraphTransport}
\end{figure}

\begin{figure}
\centering
\includegraphics[width=12cm]{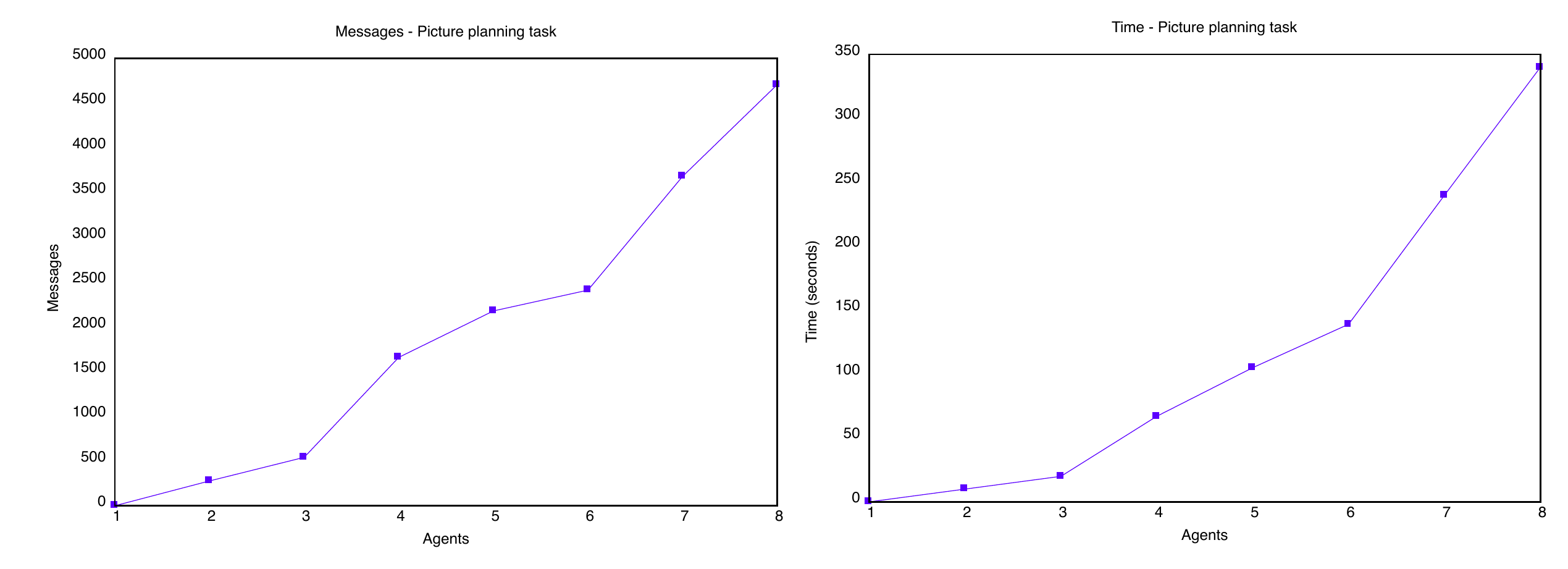}
\caption{Scalability results for the picture domain}
\label{GraphPicture}
\end{figure}
 \subsubsection{Scalability analysis}

In this subsection we evaluate the scalability of our MAP framework, i.e., how the number of agents in the MAP system affects its efficiency. To do so, eight different test problems were generated for both the \textit{transportation} and the \textit{picture} domains. Each test increases the number of agents by one, keeping the rest of the planning problem's parameters unchanged.

All the \textit{transportation} tests include ten different cities, one truck, one empty table in the warehouse and one package of raw material. All the problems include a single warehouse agent, and each of them adds an extra transport agent, up to eight transport agents. The problem goal for all the test problems is to deliver the raw material to the warehouse, which must place it on the empty table. The optimal solution plan for all the problems includes ten actions and involves the participation of at least one transport agent and the warehouse agent.

As for the \textit{picture} domain, all the test problems include two different tools and twelve different locations. The goal for all the problems is to hang two different pictures. The optimal solution plan for these problems has eight actions and involves the participation of two different agents. Each agent picks up one tool and hangs one picture.

Figures \ref{GraphTransport} and \ref{GraphPicture} depict the results for each domain. As it can be observed, the number of messages experiences a notable increase with each new agent included in the MAP process. So does the execution time, which is conditioned by the number of messages exchanged among agents.
\vspace{-0.2cm}
\paragraph{Discussion on the results.}

These results are caused by the growing number of refinement plans proposed by the agents. Refinement plans are communicated to all the agents in the MAP system, reason why the addition of a planning agent represents such an overhead as each new agent proposes and communicates a number of extra refinement plans. In addition, the refinement plans proposed by each new planning agent increase the complexity of the search tree as they may also be adopted as base plans at some point.

Notice that the number of messages is much larger in the case of the \emph{picture} problems, even though we have defined similar size and complexity problems for the two planning domains. This is due to the \emph{loosely-coupled} nature of the \emph{picture} problems because agents in this domain share the same abilities and every agent can make a plan proposal over any base plan.

As opposite to the \emph{picture} domain, agents in the \emph{transportation} domain are specialized, which makes them unable to make plan refinements over every base plan. Transport agents are limited by their working areas, while warehouse agents cannot take part in the transportation of the packages. This fact limits the number of exchanged messages, which also benefits the execution time. This way, our system proves to be more stable when solving \emph{strongly-related} problems like the \emph{transportation} tests since the addition of a new agent causes a lower increase in the number of messages, which directly affects the execution time.

In conclusion, the number of agents in the MAP system is a parameter that has a significant influence on its efficiency because the number of messages among agents constitutes one of the bottlenecks of the system. This issue is more noticeable when dealing with \emph{loosely-coupled} problems, as agents can devise plan proposals over almost any base plan, whereas our MAP system shows a more robust behavior when solving \emph{strongly-related} problems. Therefore, our immediate challenge is to reduce the number of messages between agents. This way, we will improve the scalability of the system and we will be able to test more complex planning problems.
\vspace{-0.1cm}
\section{Conclusions}
\label{Conclusions}

This article presents a MAP model that allows agents to plan under \emph{incomplete information}. Our approach is suitable to solve a wide range of MAP problems, from \emph{strongly-related} problems with a high degree of interaction among agents to simpler \emph{loosely-coupled} problems, which present limited interactions among agents. Our model allows for heterogeneous agents with different information, capabilities and private goals to cooperatively build a joint plan while handling an incomplete view of the MAP task. Agents keep their private data and share only the relevant information for their interactions with other agents, thus being unaware of part of the information managed by the rest of agents.

Shareable information is defined through our MAP language, extended from \emph{PDDL3.1}. The information exchange is carried out through the construction of a distributed Relaxed Planning Graph, by which agents share the public fluents and estimate the best cost to achieve them.

The MAP resolution process is based on a refinement planning procedure whereby agents propose successive refinements to an initially empty base plan until a consistent joint plan is obtained. This procedure, that iteratively combines planning and coordination, uses single-agent planning technology to build the refinement plans. More precisely, we adapt the POP paradigm to a MAP context, which allows agents to build refinement plans leaving details unresolved that will be gradually completed by other agents until a solution plan is found.

Conclusions drawn from the experiments show that MAP agents obtain solution plans of equal or better quality than a single-agent approach for both \emph{loosely-coupled} and \emph{strongly-related} problems. Despite agents do not have a complete view of the MAP task and keep private information, the quality of the MAP solution plans is not affected, neither in terms of number of actions nor plan duration. Hence, we can affirm that our model tackles large MAP tasks in which information is distributed among a number of planning entities at least as effectively as a single-agent planning approach working under complete information.

Moreover, our MAP approach enforces cooperation among agents since they work together to solve goals more efficiently. MAP improves plan concurrency as agents can solve different goals in parallel, which reduces the duration and the number of actions of the solution plans.

\begin{acknowledgements}
This work has been partly supported by the Spanish MICINN under projects Consolider Ingenio 2010 CSD2007-00022 and TIN2011-27652-C03-01, and the Valencian Prometeo project 2008/051.
\end{acknowledgements}

\bibliographystyle{agsm}
\bibliography{biblio}

\end{document}